\PassOptionsToPackage{table,xcdraw}{xcolor}
\documentclass[11pt, a4paper, logo, copyright, nonumbering]{wechat}
\usepackage[numbers, square, sort&compress]{natbib}
\usepackage{dblfloatfix}
\usepackage{ulem}
\usepackage{caption}
\definecolor{citecolor}{HTML}{1976D2}
\hypersetup{
    colorlinks=true,
    linkcolor=black,
    filecolor=magenta,
    urlcolor=blue!50!black,
    citecolor=citecolor,
}
\usepackage{dramatist}
\usepackage{xspace}
\usepackage{pifont}
\usepackage{multirow}
\usepackage{tcolorbox}
\usepackage{xltabular}
\usepackage{longtable}
\usepackage{hyperref}
\usepackage{diagbox}
\usepackage{makecell}
\usepackage{amsmath}
\usepackage{amssymb}
\usepackage{amsfonts}
\usepackage{bm}
\usepackage{lineno}

\usepackage[bottom]{footmisc}

\usepackage{CJKutf8}
\usepackage{subfigure}
\usepackage{setspace}
\usepackage{subcaption} 

\usepackage{titlesec}

\usepackage{tcolorbox}
\usepackage{fontawesome5}
\usepackage{enumitem}

\usepackage{booktabs}
\usepackage{multirow}
\usepackage[table,xcdraw]{xcolor}
\usepackage{graphicx}

\usepackage{algorithm}
\usepackage{algorithmic}
\usepackage{subcaption}

\tcbset{
    contributionbox/.style={
        colback=blue!3,
        colframe=blue!50!black,
        boxrule=0.5pt,
        arc=3pt,
        left=6pt,
        right=6pt,
        top=6pt,
        bottom=6pt,
        fonttitle=\bfseries
    }
}

\titlespacing*{\paragraph}
  {0pt}                   
  {0.5ex plus 1ex minus .2ex} 
  {1em}

\makeatletter
\def\@BTrule[#1]{%
  \ifx\longtable\undefined
    \let\@BTswitch\@BTnormal
  \else\ifx\hline\LT@hline
    \nobreak
    \let\@BTswitch\@BLTrule
  \else
     \let\@BTswitch\@BTnormal
  \fi\fi
  \global\@thisrulewidth=#1\relax
  \ifnum\@thisruleclass=\tw@\vskip\@aboverulesep\else
  \ifnum\@lastruleclass=\z@\vskip\@aboverulesep\else
  \ifnum\@lastruleclass=\@ne\vskip\doublerulesep\fi\fi\fi
  \@BTswitch}
\makeatother

\addto\extrasenglish{
}

 {\begin{list}{}%
         {\setlength{\leftmargin}{#1}}%
         \item[]%
 }
 {\end{list}}

\reportnumber{001} 

\title{\centering FlashPrefill V2: Block-Sparse Prefill Attention for Long-Context LLM Serving}

\author{
    \vspace{0.5em}
    {\large \bfseries
    Qihang Fan$^{1,2,3,\ast}$, Huaibo Huang$^{1, 2,\dagger}$, Zhiying Wu$^{3}$, 
    } \\ \vspace{0.6pt}
    {\large \bfseries
    Bingning Wang$^{3,\ddagger}$, Ran He$^{1, 2}$
    } \\ \vspace{1.0em}
    
    {\normalsize \normalfont
    $^{1}$MAIS\&NLPR, CASIA \quad
    $^{2}$UCAS \quad
    $^{3}$WeChat, Tencent 
    }
}

\renewcommand{\phi}{\varphi}

\renewcommand{\epsilon}{\varepsilon}
\renewcommand{\imath}{\mathrm{i}}

\newlength{\restsubwidth}
\newlength{\restsubheight}
\newlength{\restsubmoreheight}
\newcommand{\rest}[2]{%
        \settowidth{\restsubwidth}{\ensuremath{#2}}
        \settoheight{\restsubheight}{\ensuremath{{}_{#2}}}
        \ensuremath{{#1\hskip 0.5pt}_{\vrule\kern2pt\parbox[b][%
        4pt][b]{\the\restsubwidth}{%
                        \ensuremath{{}_{#2}}}}}
        }

\github{https://github.com/qhfan/FlashPrefillv2}

\begin{abstract}
Long-context modeling is a pivotal capability for Large Language Models, yet the quadratic complexity of attention remains a critical bottleneck, particularly during the compute-intensive prefilling phase. Our previous work, FlashPrefill~\cite{flashprefillv1}, mitigates this cost through instantaneous pattern discovery and max-based dynamic thresholding; however, it remains an algorithmic prototype that is still distant from production deployment. In this paper, we present FlashPrefill V2, which evolves FlashPrefill from a prototype toward practical long-context serving along three dimensions. First, we introduce a mean correction term that effectively suppresses the approximation error, keeping performance degradation manageable even at extreme sparsity levels. Second, we redesign the sparse attention operator with PackGQA memory access, warp specialization, and pingpong pipelining, fully aligning with the latest FlashAttention-3/4 implementations~\cite{flashattention3,flashattention4} and supporting FP8 inference to meet practical quantization requirements. Third, FlashPrefill V2 natively supports paged KV cache and continuous batching, allowing integration as an attention backend in modern inference frameworks such as SGLang~\cite{sglang}. Extensive evaluations on NVIDIA H20 GPUs---among the most widely deployed inference accelerators---demonstrate that FlashPrefill V2 delivers up to $\mathbf{47.26\times}$ and $\mathbf{27.19\times}$ speedups over FlashAttention-2 at 128K context length under FP8 and BF16 precision, respectively, and, in FP8, still achieves a $\mathbf{30.49\times}$ speedup against an FA3/4-aligned dense baseline.
\end{abstract}

\begin{document}
\begin{CJK*}{UTF8}{gbsn}

\maketitle

\enlargethispage{1cm}

\newcommand\blfootnote[1]{%
  \begingroup
  \renewcommand\thefootnote{}\footnote{#1}%
  \addtocounter{footnote}{-1}%
  \endgroup
}

\blfootnote{$^\dagger$ Corresponding Author.}
\blfootnote{$^\ddagger$ Project Leader.}
\blfootnote{$^\ast$ Work done during internship at WeChat.}

\section{Introduction}

Recent years have witnessed the rapid evolution of Large Language Models (LLMs)~\cite{llama,llama2,llama3}. However, the quadratic complexity of self-attention incurs prohibitive overhead on long-context sequences~\cite{attention}, a bottleneck that is particularly salient during the compute-intensive prefill stage. To mitigate this cost, various sparse attention mechanisms have been proposed~\cite{minference,flexprefill,xattention}: they coarsely estimate attention scores to identify salient blocks with Top-$k$/Top-$p$ selection, and restrict fine-grained computation to these critical tokens.

Our previous work, FlashPrefill~\cite{flashprefillv1}, pushes this line of research further with instantaneous pattern discovery and max-based dynamic thresholding, eliminating the sorting latency of Top-$k$/Top-$p$ selection. Nevertheless, it remains an algorithmic prototype that is still distant from production deployment: its accuracy can degrade uncontrollably under aggressive sparsity; its kernel is built upon FlashAttention-2~\cite{flashattention2} and lags behind state-of-the-art dense implementations such as FlashAttention-3/4~\cite{flashattention3,flashattention4}, a gap that is particularly pronounced on Hopper-architecture GPUs where TMA and asynchronous pipelines dominate kernel efficiency; and its contiguous KV layout is incompatible with the paged KV cache and continuous batching that underpin modern serving frameworks~\cite{vllm,sglang}.

\begin{figure*}[t]
    \centering
    \includegraphics[width=0.98\linewidth]{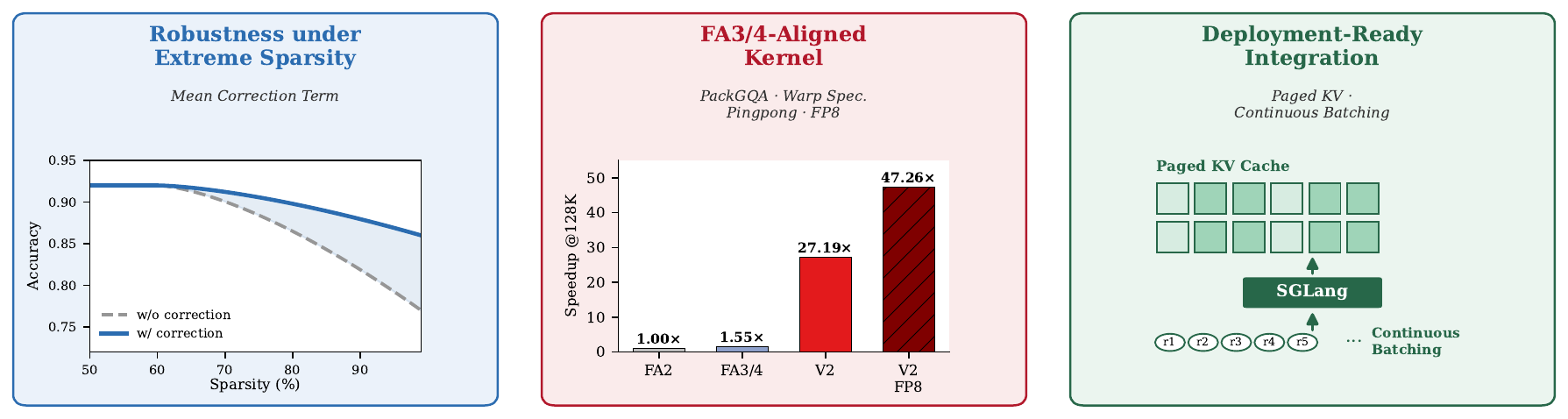}
    \caption{FlashPrefill V2 evolves FlashPrefill~\cite{flashprefillv1} from an algorithmic prototype toward practical long-context serving along three dimensions: (1) a mean correction term that preserves model accuracy under extreme sparsity; (2) an FA3/4-aligned sparse attention kernel with PackGQA memory access, warp specialization, pingpong pipelining, and FP8 support; and (3) native compatibility with paged KV cache and continuous batching for integration as an attention backend in modern serving frameworks such as SGLang~\cite{sglang}.}
    \label{fig:teaser}
\end{figure*}

In this paper, we present FlashPrefill V2, which closes these gaps as illustrated in Fig.~\ref{fig:teaser}. \textbf{(1)} We introduce a mean correction term that compensates pruned blocks with their pooled K/V statistics inside the attention computation, keeping accuracy degradation manageable even at extreme sparsity levels. \textbf{(2)} We redesign the sparse attention operator for Hopper-architecture GPUs with PackGQA memory access, warp specialization, and pingpong pipelining, fully aligning with the latest FlashAttention-3/4 implementations and supporting FP8 precision for quantized deployment. \textbf{(3)} FlashPrefill V2 natively supports paged KV cache and continuous batching, and can be integrated as an attention backend in modern inference frameworks such as SGLang.

\begin{figure*}[t]
    \centering
    \includegraphics[width=0.98\linewidth]{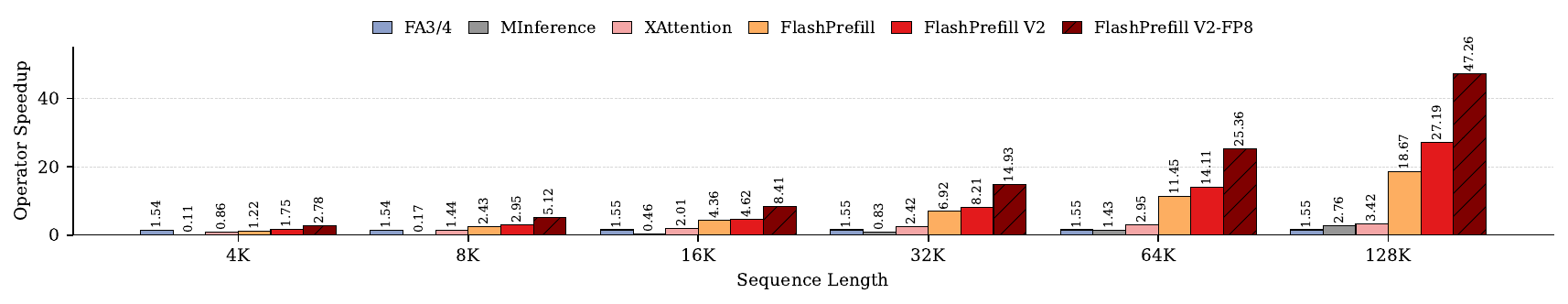}
    \caption{Speedup of various attention operators relative to FlashAttention-2~\cite{flashattention2} on NVIDIA H20 GPUs, including an FA3/4-aligned dense baseline~\cite{flashattention3,flashattention4}. All results are measured with a batch size of 4. FlashPrefill V2 exhibits a dominant advantage, particularly in long-context scenarios, and its FP8 variant further amplifies the gains.}
    \label{fig:attention_speedup}
\end{figure*}

As shown in Fig.~\ref{fig:attention_speedup}, on NVIDIA H20, one of the most widely deployed inference GPUs and a representative of the Hopper architecture, FlashPrefill V2 attains a $\mathbf{27.19\times}$ speedup over FlashAttention-2 at a 128K context length under BF16 precision, surpassing all competing sparse operators, while its FP8 variant further reaches $\mathbf{47.26\times}$. Even against the much stronger FA3/4-aligned dense baseline, FlashPrefill V2 still delivers speedups of $\mathbf{17.54\times}$ (BF16) and $\mathbf{30.49\times}$ (FP8), and maintains acceleration at sequence lengths as short as 4K.

Our contributions can be summarized as follows:
\begin{itemize}
    \item A mean correction term for block-wise score estimation that ensures controlled accuracy degradation even under extreme sparsity.
    \item A redesigned sparse attention operator featuring PackGQA memory access, warp specialization, and pingpong pipelining, fully aligned with FlashAttention-3/4 and extended to FP8 inference, attaining up to $\mathbf{47.26\times}$ speedup over FlashAttention-2 at 128K context.
    \item Native compatibility with paged KV cache and continuous batching, demonstrated through integration into SGLang with up to $4.8\times$ lower end-to-end time-to-first-token at 128K with marginal accuracy loss.
\end{itemize}

\section{Related Works}

\paragraph{Sparse Attention.}Reducing the quadratic cost of attention via sparsity has been extensively studied. Early designs fix the sparse layout in advance, via local windows with global tokens~\cite{beltagy2020longformer,bigbird} or hashing- and clustering-based routing~\cite{reformer,vyas2020clustered}, or evict KV entries under a preset budget~\cite{streamingllm,zhang2023h2o,snapkv,pyramidkv,infllm}. Learnable variants integrate sparsity into training or fine-tuning, coupling the pattern with the weights~\cite{native-sparse-attention,zhao2025infllmv2,mobamixtureblockattention,optimizingmixtureblockattention}. Training-free methods instead estimate which regions of the attention map are worth computing and skip the rest at inference~\cite{minference,flexprefill,xattention}, extend to query-aware KV selection during decoding~\cite{quest,sparq,tactic,duoattention,triforce}, and to multimodal inputs~\cite{mminference}. A recent variant compensates pruned blocks with surrogate contributions~\cite{solattn}, and UniPrefill~\cite{uniprefill} lifts block-wise sparsification to token-level computation for hybrid architectures. FlashPrefill V2 belongs to the training-free category: compared with prior training-free methods, it further emphasizes controlled accuracy under extreme sparsity and system-level deployability, rather than treating the sparse pattern as the sole design axis. A complementary line replaces softmax attention altogether with linear or recurrent formulations~\cite{retnet,rwkv7}, e.g., rectifying its neglect of query magnitude~\cite{mala}.

\paragraph{Efficient Attention Kernels.}Orthogonal to sparsity, a parallel line of work accelerates dense attention through careful hardware-aware kernel design. FlashAttention-2 reformulates the computation around better parallelism and work partitioning~\cite{flashattention2}, while FlashAttention-3 exploits asynchrony and low-precision arithmetic on Hopper GPUs~\cite{flashattention3}, and FlashAttention-4 co-designs the algorithm with kernel pipelining to accommodate asymmetric hardware scaling~\cite{flashattention4}. Our sparse operator inherits this design philosophy: built upon CUTLASS/CuTe with TMA and warp-specialized producer-consumer pipelines, it adopts PackGQA memory access so that block sparsity translates into wall-clock speedup rather than being offset by kernel inefficiency, and it further supports FP8 execution.

\paragraph{LLM Serving Systems.}Production-grade inference increasingly relies on system-level innovations in memory management and scheduling. vLLM popularized paged KV cache management with PagedAttention~\cite{vllm}, SGLang optimizes the execution of structured language model programs~\cite{sglang}, and disaggregated architectures further separate prefill from decoding to satisfy latency objectives~\cite{distserve,mooncake}; continuous batching across concurrent requests has since become a standard technique for maximizing throughput. However, most research-oriented sparse attention kernels assume a contiguous KV layout, which prevents direct adoption in these frameworks. FlashPrefill V2 is instead designed around paged KV cache and continuous batching from the outset, and can be integrated as an attention backend in modern serving engines.

\section{Method}

\begin{figure*}[t]
    \centering
    \includegraphics[width=0.98\linewidth]{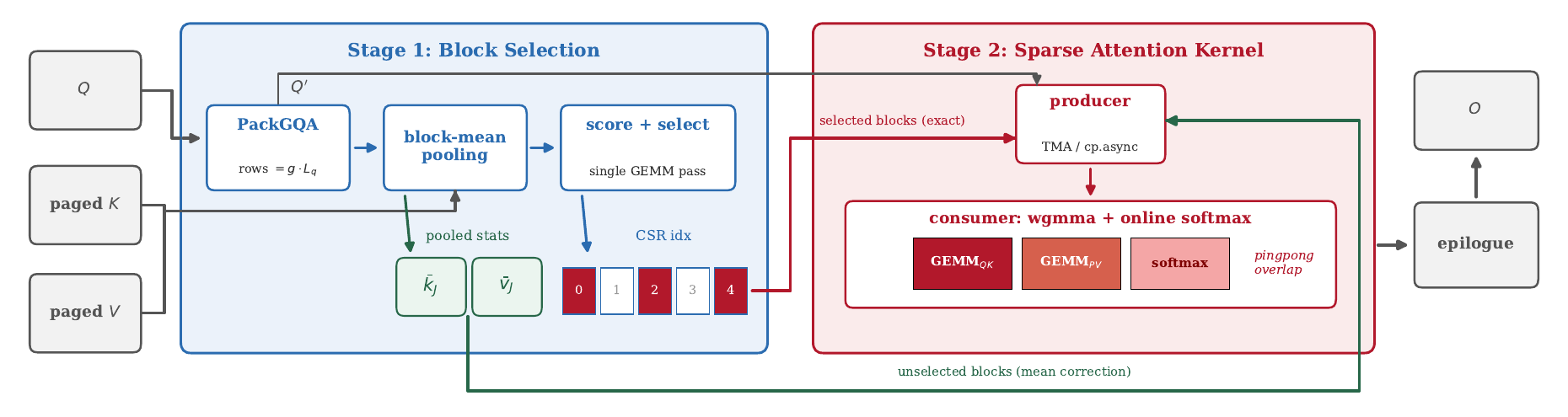}
    \caption{The FlashPrefill V2 prefill pipeline. Stage 1 packs queries with PackGQA, pools block-mean statistics, and produces a CSR index of selected blocks in a single fused pass; Stage 2 runs the warp-specialized sparse attention kernel over the selected blocks (exact path) while the unselected blocks are compensated through the mean correction path (Sec.~\ref{sec:correction}), and the two streams merge inside the online softmax.}
    \label{fig:pipeline}
\end{figure*}

\subsection{Preliminaries: The FlashPrefill Framework}
\label{sec:preliminary}
We first recap the core computation of FlashPrefill~\cite{flashprefillv1}, upon which FlashPrefill V2 is built. Consider the prefilling stage with $L$ tokens, where $Q, K, V \in \mathbb{R}^{L \times d}$ denote the query, key, and value matrices of one attention head, and the causal output is $O = \mathrm{softmax}(QK^{\top}/\sqrt{d} + \mathcal{M})\,V$. FlashPrefill partitions $K$ and $V$ into blocks of size $B$ and accelerates prefilling in two stages.

\paragraph{Block-level Score Estimation.}FlashPrefill identifies the salient blocks with uniformly distributed probe queries, which simultaneously resolve vertical, slash, and block-sparse patterns. Each key block $\mathcal{B}_J$ is represented by its mean-pooled key $\bar{k}_J = \frac{1}{B}\sum_{k_i \in \mathcal{B}_J} k_i$. Since $\exp(q \cdot \bar{k}_J)$ is the geometric mean of the per-token scores $\{\exp(q \cdot k_i)\}$, by the AM-GM inequality it lower-bounds the block's true contribution $\frac{1}{B}\sum_i \exp(q \cdot k_i)$, while the low intra-block variance of attention distributions keeps the cross-block ranking preserved. To avoid materializing the $L \times (L/B)$ score matrix, a fused kernel computes, for each query tile $I$ and key block $J$,
\begin{equation}
m_{I,J} = \max_{q_i \in I}\, (q_i \cdot \bar{k}_J), \qquad
\mathcal{S}_{I,J} = \sum_{q_i \in I} \exp(q_i \cdot \bar{k}_J - m_{I,J}),
\end{equation}
followed by a global rescaling with $\mathcal{M}_I = \max_J m_{I,J}$ and row normalization:
\begin{equation}
\mathrm{Score}_{I,J} = \frac{\mathcal{S}_{I,J} \cdot \exp(m_{I,J} - \mathcal{M}_I)}{\sum_{J'} \mathcal{S}_{I,J'} \cdot \exp(m_{I,J'} - \mathcal{M}_I) + \epsilon},
\end{equation}
reducing the memory footprint from $O(L^2/B)$ to $O((L/B)^2)$.

\paragraph{Max-based Dynamic Thresholding.}Instead of Top-$k$/Top-$p$ selection, which requires global sorting or cumulative summation, FlashPrefill derives the pruning threshold for each query block $I$ from its peak score,
\begin{equation}
\mathrm{thresh}_I = \alpha \cdot \max_{J} \mathrm{Score}_{I,J},
\end{equation}
and retains every key block with $\mathrm{Score}_{I,J} \ge \mathrm{thresh}_I$, together with the first sink blocks, a local sliding-window band, and the most recent blocks, subject to the causal constraint. Block-sparse attention is then executed only over the selected blocks.

\subsection{Mean Correction under Extreme Sparsity}
\label{sec:correction}
\begin{figure*}[t]
    \centering
    \includegraphics[width=0.98\linewidth]{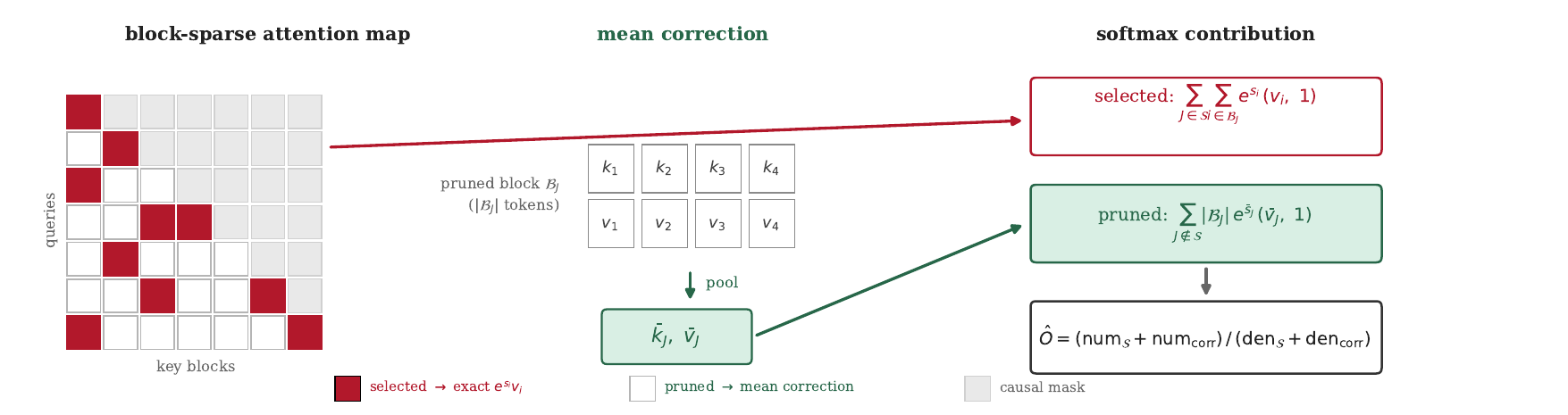}
    \caption{Mean correction. Selected blocks (red) are computed exactly, while each pruned block is pooled into its mean statistics $(\bar{k}_J, \bar{v}_J)$ and contributes a surrogate term $|\mathcal{B}_J| e^{\bar{s}_J}(\bar{v}_J, 1)$ to the softmax numerator and denominator, recovering the discarded probability mass without per-token computation.}
    \label{fig:correction}
\end{figure*}

For a query $q$ with per-token logits $s_i = q \cdot k_i / \sqrt{d}$, let $\mathcal{S}$ denote the selected block set. FlashPrefill approximates the exact output by truncating both sums of the softmax to $\mathcal{S}$:
\begin{equation}
O = \frac{\sum_{J}\sum_{i \in \mathcal{B}_J} e^{s_i} v_i}{\sum_{J}\sum_{i \in \mathcal{B}_J} e^{s_i}}
\;\;\approx\;\;
\frac{\sum_{J \in \mathcal{S}}\sum_{i \in \mathcal{B}_J} e^{s_i} v_i}{\sum_{J \in \mathcal{S}}\sum_{i \in \mathcal{B}_J} e^{s_i}}.
\end{equation}
Under extreme sparsity, however, the discarded mass $\sum_{J \notin \mathcal{S}}\sum_{i \in \mathcal{B}_J} e^{s_i}$ can no longer be neglected. We recover it with a zero-order mean correction (a surrogate-contribution idea also adopted in~\cite{solattn}), as illustrated in Fig.~\ref{fig:correction}: for each key block $J$, we precompute
\begin{equation}
\bar{k}_J = \frac{1}{|\mathcal{B}_J|} \sum_{i \in \mathcal{B}_J} k_i, \qquad
\bar{v}_J = \frac{1}{|\mathcal{B}_J|} \sum_{i \in \mathcal{B}_J} v_i, \qquad
\bar{s}_J = q \cdot \bar{k}_J / \sqrt{d},
\end{equation}
and the corrected output reads
\begin{equation}
\label{eq:correction}
\hat{O} = \frac{\displaystyle\sum_{J \in \mathcal{S}} \sum_{i \in \mathcal{B}_J} e^{s_i} v_i \;+\; \sum_{J \notin \mathcal{S}} |\mathcal{B}_J| \, e^{\bar{s}_J} \, \bar{v}_J}{\displaystyle\sum_{J \in \mathcal{S}} \sum_{i \in \mathcal{B}_J} e^{s_i} \;+\; \sum_{J \notin \mathcal{S}} |\mathcal{B}_J| \, e^{\bar{s}_J}}.
\end{equation}

\paragraph{Approximation error.}Writing $\delta_i = s_i - \bar{s}_J$ with $\sum_i \delta_i = 0$ and denoting within-block averages by $\overline{(\cdot)}$, the exact mass and weighted value of block $J$ expand around $\bar{s}_J$ as
\begin{equation}
\label{eq:taylor}
\sum_{i \in \mathcal{B}_J} e^{s_i}
= |\mathcal{B}_J| \, e^{\bar{s}_J}\Big(1 + \tfrac{\overline{\delta^2}}{2} + O(\overline{\delta^3})\Big),
\qquad
\sum_{i \in \mathcal{B}_J} e^{s_i} v_i
= |\mathcal{B}_J| \, e^{\bar{s}_J}\Big(\bar{v}_J + \overline{\delta v} + O(\overline{\delta^2 v})\Big),
\end{equation}
where $\overline{\delta v} = \frac{1}{|\mathcal{B}_J|} \sum_{i} \delta_i (v_i - \bar{v}_J)$ is the within-block covariance between logit deviations and values: the mass surrogate is second-order accurate, whereas the numerator surrogate retains a first-order covariance term. With block mass $w_J = \sum_{i \in \mathcal{B}_J} e^{s_i}$, total mass $D = \sum_J w_J$, and exact block mean $\mu_J = w_J^{-1} \sum_{i \in \mathcal{B}_J} e^{s_i} v_i$, substituting the surrogate for a pruned block shifts the output by
\begin{equation}
\label{eq:error}
\Delta_J \;=\; \frac{w_J}{D}\,\big(\mu_J - \bar{v}_J\big) + O\!\Big(\frac{w_J}{D}\,\overline{\delta^2}\Big),
\qquad
\mu_J - \bar{v}_J \;=\; \overline{\delta v} + O(\overline{\delta^2 v}).
\end{equation}
Eq.~\ref{eq:error} is controlled by three bounds. (i) Max-based thresholding (Sec.~\ref{sec:preliminary}) caps the mass share of every pruned block, $w_J / D \lesssim \alpha$, so the total shift is a small-share signed sum,
\begin{equation}
\label{eq:total}
\Big\|\sum_{J \notin \mathcal{S}} \Delta_J\Big\| \;\le\; \rho_{\mathrm{pr}} \cdot \max_{J \notin \mathcal{S}} \big\|\mu_J - \bar{v}_J\big\|,
\qquad \rho_{\mathrm{pr}} = \sum_{J \notin \mathcal{S}} \frac{w_J}{D}.
\end{equation}
(ii) The per-block deviation is a covariance, vanishing under within-block independence and bounded by dispersion via Cauchy--Schwarz,
\begin{equation}
\label{eq:cov}
\overline{\delta v} = 0 \;\;\text{if}\;\; \delta \perp (v - \bar{v}_J),
\qquad
\big\|\overline{\delta v}\big\| \;\le\; \delta_{\mathrm{rms}} \cdot \sigma_v,
\end{equation}
where $\sigma_v$ is the within-block value dispersion, itself small under the low intra-block variation that justifies block-mean score estimation (Sec.~\ref{sec:preliminary}). (iii) Relative to dropping the block outright, $\Delta_J^{\mathrm{none}} = \frac{w_J}{D}(\mu_J - O)$, the correction cancels the zeroth order and retains only the covariance:
\begin{equation}
\label{eq:ratio}
\frac{\|\Delta_J\|}{\|\Delta_J^{\mathrm{none}}\|} \;=\; \frac{\|\mu_J - \bar{v}_J\|}{\|\mu_J - O\|} \;\sim\; O(\delta_{\mathrm{rms}}),
\end{equation}
since $\mu_J - O$ is an inter-block deviation while $\mu_J - \bar{v}_J$ is an intra-block one. Salient blocks always follow the exact path, so only the small-$w_J$ tail enters Eq.~\ref{eq:total}. Empirically (Tab.~\ref{tab:ablation_correction}), removing the correction costs up to $6.2$ points at 128K in FP8, while the corrected pipeline stays within $1.5$ points of full attention even at 128K in BF16.

\paragraph{Implementation.}As illustrated by the green correction path in Fig.~\ref{fig:pipeline}, only blocks fully visible to the whole query tile are corrected; blocks intersecting the diagonal band are covered by the exact sink/window/recent selections. Within the kernel, each corrected block enters the pipeline as one extra iteration whose score is shifted by $\log |\mathcal{B}_J|$, so that a single mean vector represents $|\mathcal{B}_J|$ tokens in both the numerator and the denominator of Eq.~\ref{eq:correction}, reusing the same MMA pipeline and online-softmax rescaling without any additional kernel launch. Concretely, a two-pointer scan over the selected-block list aggregates the correction targets of each chunk into a bitmask, $\mathrm{todo} = \mathrm{valid} \mathbin{\&} \lnot\,\mathrm{selected}$, and chunks in which every block is selected are skipped identically by the producer and the consumer, so the pipeline stays in lockstep without extra synchronization.

\subsection{Hopper-Aligned Sparse Attention Operator}
\label{sec:kernel}
The V1 sparse kernel inherits the execution model of FlashAttention-2, whose tile-based synchronous pipeline under-utilizes Hopper GPUs: the Tensor Memory Accelerator (TMA) and asynchronous warpgroup MMAs (wgmma) only reach full throughput when data movement and computation are decoupled. We therefore redesign the operator on top of a CUTLASS/CuTe warp-specialized pipeline (Stage 2 of Fig.~\ref{fig:pipeline}). Our implementation builds upon the Hopper-specific techniques established by FlashAttention-3/4 themselves~\cite{flashattention3,flashattention4}---TMA-based asynchronous data movement, warp-specialized producer-consumer pipelines, and intra-warpgroup GEMM-softmax overlap---and our contribution lies in extending this execution model to block-sparse computation (index-driven traversal, correction chunk injection) and to the GQA/paged/FP8 regimes that deployment requires, as detailed below.

\paragraph{PackGQA Memory Layout.}Under grouped-query attention with ratio $g = H_q / H_{kv}$, a naive mapping assigns one thread block to each (query head, $M$-tile) pair, so that $g$ thread blocks repeatedly load the same KV block. We instead reshape the query matrix as
\begin{equation}
Q \in \mathbb{R}^{L \times H_q \times d} \;\longrightarrow\; Q' \in \mathbb{R}^{(g \cdot L) \times H_{kv} \times d},
\end{equation}
where the packed row index $r$ maps to position and head via
\begin{equation}
\label{eq:packgqa}
\text{pos}(r) = \lfloor r / g \rfloor, \qquad h(r) = h_{kv} \cdot g + (r \bmod g),
\end{equation}
implemented with a single fast divmod per row. Each $M$-tile of $Q'$ covers $kBlockM/g$ positions of all $g$ heads of one KV group, yielding three properties: (i) each staged KV block is consumed by all rows of the tile, so group-wide sharing is guaranteed in shared memory and never relies on cross-CTA L2 retention; (ii) the sparse index of Sec.~\ref{sec:preliminary} is defined per KV head, shrinking the index metadata by a factor of $g$; and (iii) tiles stay fully occupied for any $L$: the wasted-row fraction is $(\lceil L/kBlockM \rceil kBlockM - L)/(\lceil L/kBlockM \rceil kBlockM)$ per head under the naive mapping but only $(\lceil gL/kBlockM \rceil kBlockM - gL)/(\lceil gL/kBlockM \rceil kBlockM)$ once per group under packing, which for $L \le kBlockM/g$ (e.g., decode) cuts KV stagings by up to the full factor of $g$. Q is staged with vectorized \texttt{cp.async} loads, with threads of the same row sharing one warp to amortize page-table lookups.

\paragraph{Warp-Specialized Producer-Consumer Pipeline.}The persistent kernel partitions its warpgroups into one producer warpgroup issuing asynchronous TMA/cp.async loads of K/V (and of the correction chunks of Sec.~\ref{sec:correction}) into a multi-stage shared-memory pipeline, and two consumer warpgroups executing wgmma on staged tiles, with registers reallocated between the two roles (\texttt{warpgroup\_reg\_dealloc/alloc}). On top of this, GEMM and softmax proceed in a pingpong fashion within each consumer warpgroup: each step issues the QK GEMM of block $n$, the PV GEMM of block $n{+}1$, and the online softmax of block $n$,
\begin{equation}
\label{eq:pingpong}
\underbrace{S^{(n)} = Q\,K_{(n)}^{\top}}_{\text{GEMM}_0} \;\parallel\; \underbrace{O \mathrel{+}= P^{(n+1)}\,V_{(n+1)}}_{\text{GEMM}_1} \;\parallel\; \underbrace{\tilde{P}^{(n)} = \mathrm{online\text{-}softmax}(S^{(n)})}_{\text{softmax}},
\end{equation}
so that MMA and non-MMA (exponentiation, rescaling) units overlap instead of stalling each other. This matches the intra-warpgroup overlap schedule of state-of-the-art dense implementations~\cite{flashattention3,flashattention4}, which our dense path replicates to provide a strong baseline.

\paragraph{FP8 Execution.}With per-tensor scales $c_q, c_k, c_v$, FP8-e4m3 operands are dequantized on the fly inside the online softmax:
\begin{equation}
\label{eq:fp8}
S = c_q c_k \cdot \big(\tilde{Q}\,\tilde{K}^{\top}\big), \qquad
\tilde{P} = \big\lfloor 2^{8} \cdot \exp(S - m) \big\rceil_{\mathrm{e4m3}} \in [0, 256], \qquad
O = c_v \cdot \frac{\sum_t \tilde{P}^{(t)} \tilde{V}^{(t)}}{\sum_t \tilde{P}^{(t)}},
\end{equation}
where $\lfloor \cdot \rceil_{\mathrm{e4m3}}$ denotes FP8 quantization and $m$ the running row maximum. The $2^{8}$ offset maps probabilities onto $[0, 256]$, fully utilizing the e4m3 dynamic range; it cancels in the softmax ratio and is recovered only in the log-sum-exp at finalize.

The second GEMM requires special handling because FP8 wgmma mandates a K-major (transposed) layout for the B operand, which constrains how $\tilde{V}$ is fed:
\begin{itemize}
    \item \textbf{Column-major $\tilde{V}$} already satisfies the layout and is consumed with zero copying; the probability fragment $\tilde{P}$ is instead re-arranged into the wgmma A-operand register layout purely in registers via warp shuffles and byte permutation, and the output columns are un-permuted symmetrically in the epilogue:
    \begin{equation}
    \label{eq:permute}
    \tilde{P} \xrightarrow{\ \mathrm{shfl}+\mathrm{byte\_perm}\ } \tilde{P}', \qquad
    O' = \tilde{P}'\,\tilde{V} \xrightarrow{\ \mathrm{epilogue}\ } O.
    \end{equation}
    \item \textbf{Row-major $\tilde{V}$} is transposed on the fly by the producer inside shared memory with an LDSM.T$\to$byte-permute$\to$STSM sequence; to avoid bank conflicts we permute the columns of $\tilde{V}$ during this transpose and restore the column order of $O$ in the epilogue, rather than transposing conflict-free at full price.
\end{itemize}
In both cases the layout conversions involve no global-memory traffic, and the PV GEMM consumes $\tilde{P}$ directly from registers (RS-mode wgmma).

\paragraph{Index-Driven Sparse Traversal.}The selection result is stored as a CSR index: for each (batch, head, tile) segment $g$ with offsets $\mathrm{cu}[g]$ and $T = \mathrm{cu}[g{+}1] - \mathrm{cu}[g]$, the ascending index list $\{\mathrm{idx}[\mathrm{cu}[g] + i]\}_{i=0}^{T-1}$ defines the iterated block sequence
\begin{equation}
\label{eq:csr}
n^{(t)} = \mathrm{idx}\big[\mathrm{cu}[g] + T - 1 - t\big], \qquad t = 0, \dots, T-1,
\end{equation}
traversed backwards to match the descending order of the dense schedule. Producer and consumer evaluate the same sequence $\{n^{(t)}\}$ independently, so their control flows are identical by construction and the pipeline advances in lockstep without any synchronization for block skipping; the next index $n^{(t+1)}$ is prefetched while the load and GEMM of $n^{(t)}$ are still in flight, keeping index resolution off the critical path. Under KV splitting, split $u$ of $U$ receives the contiguous subsequence
\begin{equation}
\label{eq:split}
t \in \big[\, u \cdot \lceil T/U \rceil,\; \min\big((u{+}1) \lceil T/U \rceil,\, T\big) \,\big),
\end{equation}
i.e., the CSR list is partitioned by selected-block count rather than by position range, equalizing actual work across splits.

\paragraph{Single-Pass Score and Selection.}Scoring and thresholding are fused into one pass over the pooled keys. With $s_{iJ} = q_i \cdot \bar{k}_J / \sqrt{d}$, the scoring kernel maintains a running tile maximum $M$ and block energies online,
\begin{equation}
\label{eq:fused_score}
M \leftarrow \max(M, M_{\text{tile}}), \qquad
E_J \;=\; \sum_{q_i \in I} 2^{\;s_{iJ} \log_2 e \;-\; M},
\end{equation}
then writes $E_J$ (bitcast to integer) into the index buffer itself together with the per-chunk maximum, eliminating both a separate score buffer and a second GEMM pass. Selection then only rescales and thresholds:
\begin{equation}
\label{eq:select}
\hat{E}_J = E_J \cdot 2^{\,M_c - M_{\text{final}}}, \qquad
J \in \mathcal{S} \;\Longleftrightarrow\; \hat{E}_J \;\ge\; \alpha \cdot \max_{J'} \hat{E}_{J'} \;\lor\; J \in \{\text{sink}\} \cup \{\text{window}\} \cup \{\text{diag}\},
\end{equation}
and survivors are compacted in place, which is safe since the write position never exceeds the scan position.

\paragraph{Sparsity-Aware Load Balancing.}After PackGQA, the cost of tile $I$ is proportional to its selected-block count,
\begin{equation}
\label{eq:cost}
C(I) \;\propto\; T_I \;=\; O(L)\ \text{for the last tiles}, \qquad O(1)\ \text{otherwise},
\end{equation}
so the dense heuristic that disables splitting once tiles cover the SMs leaves the heavy tail tiles serialized. We therefore always admit splits on the block-sparse path: requests are scheduled in descending order of their causal workload by a device-side sort in the scheduler preparation pass, and each tile's selected-block list is partitioned evenly across splits by count (Eq.~\ref{eq:split}), so that every split receives an equal share of the actual sparse work; tile assignment is resolved by warp-level parallel prefix scans over per-request lengths.

\paragraph{Paged and Variable-Length Execution.}K/V addressing goes through the page table: the global address of key token $j$ of request $b$ is
\begin{equation}
\label{eq:page}
\mathrm{addr}(b, j) = \mathrm{page\_table}\big[b,\ \lfloor j / p \rfloor\big] \cdot p + (j \bmod p),
\end{equation}
with page size $p$, resolved on the fly with \texttt{cp.async}; a varlen-aware persistent scheduler enumerates $(b, h, I)$ work units from per-request lengths $(L_b^q, L_b^{kv})$, which is exactly the execution model of continuous batching. The CSR index format is compatible with SGLang, allowing direct integration as an attention backend (Sec.~\ref{sec:experiments}).

\paragraph{Implementation Details.}All instantiations share one tile configuration ($128 \times 64$, two-stage smem pipeline, $B/64$ physical tiles per logical block); registers are statically reallocated between roles: the producer holds 24 registers with TMA-based KV (40 on the cp.async paged path), and each consumer warpgroup holds 240 (232 on the paged path).

\subsection{End-to-End Prefill Pipeline}
\label{sec:pipeline}
Fig.~\ref{fig:pipeline} and Algorithm~\ref{alg:pipeline} put the pieces together. Given a batch of requests with new-token lengths $\{L_b^q\}$ and cached KV lengths $\{L_b^{kv}\}$, FlashPrefill V2 executes two stages per layer: an index stage (Sec.~\ref{sec:preliminary}) that packs queries, scores pooled blocks, and compacts the survivors into a CSR index, and an attention stage (Sec.~\ref{sec:correction},~\ref{sec:kernel}) that runs the warp-specialized sparse operator over the selected blocks with mean correction.

\begin{algorithm}[t]
\caption{FlashPrefill V2 prefill pipeline (one attention layer)}
\label{alg:pipeline}
\begin{algorithmic}[1]
\REQUIRE packed queries $Q$, paged KV cache $(K, V)$ with page table, lengths $\{(L_b^q, L_b^{kv})\}$, threshold $\alpha$
\ENSURE attention output $O$
\STATE $Q' \leftarrow \mathrm{PackGQA}(Q)$, with row mapping of Eq.~\ref{eq:packgqa}
\STATE $(\bar{k}_J, \bar{v}_J) \leftarrow$ block means of $(K, V)$, one Triton pass (fp8 descaled if needed)
\STATE $E_J, M \leftarrow$ fused scoring of Eq.~\ref{eq:fused_score} over $(Q', \bar{k})$ \hfill // single GEMM pass
\STATE $\mathcal{S} \leftarrow$ thresholding and in-place compaction of Eq.~\ref{eq:select}, forcing sink/window/diag blocks
\STATE $(\mathrm{cu}, \mathrm{idx}) \leftarrow$ CSR index of $\mathcal{S}$, logical blocks expanded to $B/64$ physical tiles
\FOR{each work unit $(b, h, I)$ from the persistent scheduler (descending workload order)}
  \STATE producer: stage $K, V$ tiles of $\{n^{(t)}\}$ (Eq.~\ref{eq:csr}) via TMA/cp.async through the page table (Eq.~\ref{eq:page})
  \STATE consumer: wgmma + online softmax in pingpong order (Eq.~\ref{eq:pingpong})
  \STATE correction: append mean chunks of unselected blocks with logit shift $\log|\mathcal{B}_J|$ (Eq.~\ref{eq:correction})
\ENDFOR
\STATE epilogue: rescale by the accumulated softmax denominator (and un-permute for fp8); \RETURN $O$
\end{algorithmic}
\end{algorithm}

\paragraph{Overhead Analysis.}For one KV head with group size $g$, query length $L_q$, and KV length $L$, the FLOP costs of the three components are
\begin{equation}
\label{eq:complexity}
C_{\text{idx}} = O\!\Big(g L_q \cdot \tfrac{L}{B} \cdot d\Big), \qquad
C_{\text{corr}} = O\!\Big(g L_q \cdot \tfrac{(1-\rho)L}{B} \cdot d\Big), \qquad
C_{\text{attn}} = O\!\big(g L_q \cdot \rho L \cdot 2d\big),
\end{equation}
where $\rho$ is the retained density, $B$ the block size, and the factor of two accounts for the QK and PV GEMMs. Both the index stage and the correction operate on one pooled vector per block, hence a factor of $B$ below the dense reference; relative to the sparse attention itself,
\begin{equation}
\frac{C_{\text{idx}} + C_{\text{corr}}}{C_{\text{attn}}} = O\!\Big(\frac{2 - \rho}{2 \rho \, B}\Big),
\end{equation}
i.e., the auxiliary FLOPs vanish as the retained density grows, and even at $\rho = 4\%$ with $B = 128$ they are bounded by roughly $19\%$, consistent with the gap between the attention-only and end-to-end speedups at long contexts in Fig.~\ref{fig:attention_speedup}.

\paragraph{Serving Integration.}FlashPrefill V2 is integrated into SGLang as a standard attention backend. The two-stage pipeline runs in the extend (prefill) phase, while decode falls back to dense attention, since a single query token per step leaves no room for block-level sparsity. The page-level index required by our kernels is derived from SGLang's token-level mapping by striding every $p$ entries, and per-forward metadata is built once per step and shared by all layers. Selection hyperparameters ($\alpha$, sink/window sizes) are exposed as server arguments, and the index stage reuses a per-stream workspace cache whose capacity-matched reuse avoids any host synchronization or allocator round trip in steady state. No changes to the model definition, the KV cache layout, or the scheduling logic are required.

\begin{table*}[!t]
    \centering
    \resizebox{\textwidth}{!}{
    \begin{tabular}{l|cccccc|c|cccccc}
    \toprule[1pt]
        \multirow{2}{*}{Method} & \multicolumn{7}{c|}{RULER Score} & \multicolumn{6}{c}{Operator Speedup over FA2} \\
        \cmidrule(lr){2-8} \cmidrule(lr){9-14}
        & 4K & 8K & 16K & 32K & 64K & 128K & Avg. & 4K & 8K & 16K & 32K & 64K & 128K \\
        \midrule[0.5pt]
        \multicolumn{14}{c}{\textit{Llama-3.1-8B-Instruct}} \\
        \midrule[0.5pt]
        Full & 95.98 & 94.74 & 93.07 & 89.06 & 86.27 & 73.82 & 88.82 & 1.00$\times$ & 1.00$\times$ & 1.00$\times$ & 1.00$\times$ & 1.00$\times$ & 1.00$\times$ \\
        MInference & 95.06 & 94.23 & 92.18 & 87.68 & 83.01 & 71.61 & 87.30 & 0.12$\times$ & 0.18$\times$ & 0.46$\times$ & 0.83$\times$ & 1.34$\times$ & 2.45$\times$ \\
        FlexPrefill & 95.18 & 94.09 & 92.35 & 86.79 & 84.08 & 72.01 & 87.42 & 0.12$\times$ & 0.33$\times$ & 0.98$\times$ & 2.21$\times$ & 4.16$\times$ & 5.18$\times$ \\
        XAttention & 95.24 & 94.37 & 91.61 & 87.03 & 83.65 & 72.33 & 87.37 & 0.79$\times$ & 1.29$\times$ & 1.83$\times$ & 2.34$\times$ & 3.19$\times$ & 3.48$\times$ \\
        FlashPrefill & 95.16 & 94.29 & 92.11 & 87.49 & 84.18 & 70.91 & 87.36 & 1.21$\times$ & 2.38$\times$ & 4.31$\times$ & 7.46$\times$ & 13.62$\times$ & 22.67$\times$ \\
        FlashPrefill V2 & 95.52 & 94.65 & 92.95 & 88.48 & 83.07 & 72.08 & \textbf{87.79} & \textbf{1.66$\times$} & \textbf{2.62$\times$} & \textbf{4.28$\times$} & \textbf{7.63$\times$} & \textbf{14.39$\times$} & \textbf{27.21$\times$} \\
        \textcolor{gray}{FlashPrefill V2-FP8$^*$} & \textcolor{gray}{95.28} & \textcolor{gray}{94.48} & \textcolor{gray}{92.78} & \textcolor{gray}{88.80} & \textcolor{gray}{80.28} & \textcolor{gray}{67.78} & \textcolor{gray}{\textbf{86.57}} & \textcolor{gray}{\textbf{2.67$\times$}} & \textcolor{gray}{\textbf{4.56$\times$}} & \textcolor{gray}{\textbf{7.79$\times$}} & \textcolor{gray}{\textbf{13.89$\times$}} & \textcolor{gray}{\textbf{25.76$\times$}} & \textcolor{gray}{\textbf{47.33$\times$}} \\
        \midrule[0.5pt]
        \multicolumn{14}{c}{\textit{Qwen3-4B-Instruct-2507}} \\
        \midrule[0.5pt]
        Full & 94.82 & 92.98 & 91.68 & 88.86 & 82.81 & 71.22 & 87.06 & 1.00$\times$ & 1.00$\times$ & 1.00$\times$ & 1.00$\times$ & 1.00$\times$ & 1.00$\times$ \\
        MInference & 94.06 & 92.18 & 91.09 & 88.07 & 82.08 & 68.01 & 85.92 & 0.11$\times$ & 0.15$\times$ & 0.42$\times$ & 0.83$\times$ & 1.32$\times$ & 2.53$\times$ \\
        FlexPrefill & 93.81 & 90.96 & 91.41 & 87.62 & 81.37 & 67.24 & 85.40 & 0.12$\times$ & 0.33$\times$ & 0.99$\times$ & 2.28$\times$ & 4.01$\times$ & 5.27$\times$ \\
        XAttention & 94.42 & 91.49 & 91.42 & 87.31 & 81.56 & 68.24 & 85.74 & 0.81$\times$ & 1.35$\times$ & 1.92$\times$ & 2.38$\times$ & 3.07$\times$ & 3.43$\times$ \\
        FlashPrefill & 94.71 & 92.04 & 91.28 & 87.29 & 81.08 & 69.02 & 85.90 & 1.21$\times$ & 2.40$\times$ & 4.36$\times$ & 6.91$\times$ & 11.54$\times$ & 19.26$\times$ \\
        FlashPrefill V2 & 94.49 & 92.42 & 91.56 & 87.48 & 81.67 & 69.76 & \textbf{86.23} & \textbf{1.72$\times$} & \textbf{2.82$\times$} & \textbf{4.63$\times$} & \textbf{7.92$\times$} & \textbf{13.93$\times$} & \textbf{27.45$\times$} \\
        FlashPrefill V2-FP8 & 94.13 & 92.37 & 91.36 & 87.47 & 80.59 & 68.97 & \textbf{85.82} & \textbf{2.75$\times$} & \textbf{4.87$\times$} & \textbf{8.40$\times$} & \textbf{14.44$\times$} & \textbf{25.06$\times$} & \textbf{47.59$\times$} \\
        \midrule[0.5pt]
        \multicolumn{14}{c}{\textit{Qwen3-30B-A3B-Instruct-2507}} \\
        \midrule[0.5pt]
        Full & 95.78 & 95.23 & 94.36 & 92.72 & 90.11 & 84.12 & 92.05 & 1.00$\times$ & 1.00$\times$ & 1.00$\times$ & 1.00$\times$ & 1.00$\times$ & 1.00$\times$ \\
        MInference & 95.71 & 94.33 & 93.81 & 92.01 & 89.02 & 82.09 & 91.16 & 0.11$\times$ & 0.17$\times$ & 0.46$\times$ & 0.83$\times$ & 1.43$\times$ & 2.76$\times$ \\
        FlexPrefill & 95.32 & 94.71 & 93.52 & 91.81 & 88.09 & 83.11 & 91.09 & 0.12$\times$ & 0.34$\times$ & 0.98$\times$ & 2.31$\times$ & 3.92$\times$ & 5.43$\times$ \\
        XAttention & 95.43 & 94.12 & 92.78 & 91.06 & 88.39 & 82.23 & 90.67 & 0.86$\times$ & 1.44$\times$ & 2.01$\times$ & 2.42$\times$ & 2.95$\times$ & 3.42$\times$ \\
        FlashPrefill & 95.62 & 94.47 & 93.28 & 92.17 & 88.14 & 82.96 & 91.11 & 1.22$\times$ & 2.43$\times$ & 4.36$\times$ & 6.92$\times$ & 11.45$\times$ & 18.67$\times$ \\
        FlashPrefill V2 & 95.35 & 94.85 & 94.19 & 92.48 & 89.88 & 83.83 & \textbf{91.76} & \textbf{1.75$\times$} & \textbf{2.95$\times$} & \textbf{4.62$\times$} & \textbf{8.21$\times$} & \textbf{14.11$\times$} & \textbf{27.19$\times$} \\
        FlashPrefill V2-FP8 & 95.28 & 94.53 & 94.12 & 92.34 & 89.54 & 82.55 & \textbf{91.39} & \textbf{2.78$\times$} & \textbf{5.12$\times$} & \textbf{8.41$\times$} & \textbf{14.93$\times$} & \textbf{25.36$\times$} & \textbf{47.26$\times$} \\
    \bottomrule[1pt]
    \end{tabular}}
    \caption{RULER scores (left) and attention operator speedups over FlashAttention-2 (right) at sequence lengths from 4K to 128K. The speedups are measured end-to-end on needle-in-a-haystack inputs, including pattern discovery, thresholding, and mean correction. Results marked with $^*$ (in gray) are measured with direct online FP8 quantization without corrected weights.}
    \label{tab:ruler}
\end{table*}

\begin{table*}[!t]
    \centering
    \resizebox{\textwidth}{!}{
    \begin{tabular}{l|ccc|cccc|cccc|cccc|cc|cc|cc|c}
    \toprule[1pt]
        \multirow{2}{*}{Method} & \multicolumn{3}{c|}{Single-Document QA} & \multicolumn{4}{c|}{Multi-Document QA} & \multicolumn{4}{c|}{Summarization} & \multicolumn{4}{c|}{Few-shot Learning} & \multicolumn{2}{c|}{Synthetic} & \multicolumn{2}{c|}{Code} & \multirow{2}{*}{Avg.} \\
        \cmidrule(lr){2-4} \cmidrule(lr){5-8} \cmidrule(lr){9-12} \cmidrule(lr){13-16} \cmidrule(lr){17-18} \cmidrule(lr){19-20}
        & \rotatebox{60}{NarrQA} & \rotatebox{60}{Qasper} & \rotatebox{60}{MF-en} & \rotatebox{60}{MF-zh} & \rotatebox{60}{Hotpot} & \rotatebox{60}{2Wiki} & \rotatebox{60}{MuSiQue} & \rotatebox{60}{DuRead} & \rotatebox{60}{GovRep} & \rotatebox{60}{QMSum} & \rotatebox{60}{MNews} & \rotatebox{60}{VCSum} & \rotatebox{60}{TREC} & \rotatebox{60}{TQA} & \rotatebox{60}{SAMSum} & \rotatebox{60}{LSHT} & \rotatebox{60}{Count} & \rotatebox{60}{PR-en} & \rotatebox{60}{PR-zh} & \rotatebox{60}{LCC} & \rotatebox{60}{RepB} &  \\
        \midrule[0.5pt]
        \multicolumn{23}{c}{\textit{Llama-3.1-8B-Instruct}} \\
        \midrule[0.5pt]
        Full & 29.39 & 44.61 & 56.34 & 63.63 & 57.48 & 48.08 & 32.28 & 34.43 & 34.51 & 25.29 & 26.80 & 17.37 & 72.50 & 91.48 & 43.59 & 46.50 & 11.10 & 100.00 & 90.45 & 62.93 & 56.19 & 49.76 \\
        MInference & 28.13 & 42.36 & 53.67 & 62.08 & 56.28 & 46.72 & 32.01 & 33.41 & 33.28 & 25.16 & 27.02 & 16.84 & 69.50 & 90.83 & 42.87 & 43.00 & 6.00 & 96.00 & 84.00 & 60.41 & 55.92 & 47.88 \\
        FlexPrefill & 27.62 & 42.67 & 52.98 & 62.34 & 57.14 & 45.32 & 30.97 & 33.72 & 33.87 & 24.58 & 26.34 & 17.02 & 67.50 & 90.36 & 44.72 & 43.00 & 6.00 & 94.00 & 83.33 & 61.72 & 55.83 & 47.67 \\
        XAttention & 28.52 & 43.12 & 54.01 & 60.19 & 56.73 & 45.98 & 32.17 & 33.15 & 34.55 & 25.37 & 27.21 & 17.41 & 71.00 & 91.37 & 44.05 & 44.00 & 6.88 & 95.50 & 85.68 & 61.05 & 55.36 & 48.25 \\
        FlashPrefill & 28.31 & 42.89 & 54.23 & 61.97 & 57.01 & 45.76 & 31.96 & 33.60 & 34.02 & 24.83 & 26.47 & 17.13 & 71.00 & 91.12 & 44.21 & 46.50 & 7.00 & 95.50 & 83.45 & 62.33 & 53.41 & 48.22 \\
        FlashPrefill V2 & 30.21 & 44.56 & 55.97 & 62.37 & 58.02 & 48.84 & 32.57 & 33.78 & 34.90 & 25.24 & 27.13 & 17.37 & 72.50 & 91.45 & 44.76 & 45.50 & 7.39 & 97.00 & 89.47 & 62.68 & 53.86 & \textbf{49.31} \\
        \textcolor{gray}{FlashPrefill V2-FP8$^*$} & \textcolor{gray}{29.08} & \textcolor{gray}{45.51} & \textcolor{gray}{54.70} & \textcolor{gray}{62.59} & \textcolor{gray}{55.41} & \textcolor{gray}{46.40} & \textcolor{gray}{32.03} & \textcolor{gray}{32.97} & \textcolor{gray}{34.70} & \textcolor{gray}{24.88} & \textcolor{gray}{26.97} & \textcolor{gray}{17.61} & \textcolor{gray}{70.00} & \textcolor{gray}{91.16} & \textcolor{gray}{44.80} & \textcolor{gray}{45.00} & \textcolor{gray}{6.88} & \textcolor{gray}{96.50} & \textcolor{gray}{85.68} & \textcolor{gray}{63.65} & \textcolor{gray}{57.39} & \textcolor{gray}{\textbf{48.76}} \\
        \midrule[0.5pt]
        \multicolumn{23}{c}{\textit{Qwen3-4B-Instruct-2507}} \\
        \midrule[0.5pt]
        Full & 27.97 & 44.98 & 51.12 & 64.59 & 59.39 & 44.18 & 24.85 & 26.17 & 30.68 & 22.62 & 23.93 & 11.73 & 74.00 & 87.74 & 45.55 & 44.00 & 0.75 & 100.00 & 98.04 & 62.99 & 55.56 & 47.66 \\
        MInference & 25.89 & 41.69 & 48.26 & 62.16 & 54.17 & 42.76 & 22.16 & 24.89 & 29.12 & 21.98 & 23.15 & 11.28 & 72.00 & 87.62 & 44.32 & 42.50 & 0.00 & 91.00 & 94.67 & 60.92 & 54.48 & 45.48 \\
        FlexPrefill & 25.26 & 41.48 & 48.02 & 61.98 & 54.96 & 43.21 & 22.71 & 25.43 & 30.94 & 22.15 & 24.47 & 11.47 & 70.50 & 86.18 & 45.87 & 41.00 & 0.50 & 94.00 & 94.20 & 61.67 & 56.19 & 45.82 \\
        XAttention & 24.31 & 42.67 & 49.01 & 63.02 & 54.14 & 43.14 & 22.56 & 25.12 & 29.63 & 22.41 & 23.68 & 11.56 & 72.50 & 86.47 & 45.13 & 41.50 & 0.00 & 96.00 & 93.89 & 62.24 & 54.92 & 45.90 \\
        FlashPrefill & 25.06 & 42.78 & 49.22 & 62.31 & 55.07 & 42.97 & 23.42 & 25.66 & 29.80 & 22.09 & 23.32 & 11.62 & 72.00 & 87.95 & 45.42 & 42.00 & 0.00 & 92.00 & 94.00 & 62.71 & 53.23 & 45.84 \\
        FlashPrefill V2 & 27.56 & 43.73 & 51.07 & 63.72 & 56.39 & 43.11 & 23.18 & 26.36 & 30.48 & 22.67 & 23.88 & 11.76 & 74.00 & 87.74 & 45.78 & 41.25 & 0.50 & 96.00 & 97.17 & 63.50 & 56.38 & \textbf{46.96} \\
        FlashPrefill V2-FP8 & 26.55 & 42.70 & 50.28 & 61.71 & 55.21 & 41.95 & 24.50 & 26.03 & 30.74 & 22.47 & 24.07 & 11.48 & 73.50 & 88.16 & 45.89 & 41.25 & 0.50 & 93.00 & 94.67 & 60.85 & 52.06 & \textbf{46.07} \\
        \midrule[0.5pt]
        \multicolumn{23}{c}{\textit{Qwen3-30B-A3B-Instruct-2507}} \\
        \midrule[0.5pt]
        Full & 31.29 & 44.09 & 54.79 & 66.48 & 63.86 & 56.56 & 32.29 & 24.64 & 31.10 & 21.90 & 23.48 & 11.09 & 76.50 & 91.01 & 45.79 & 51.00 & 16.50 & 100.00 & 100.00 & 72.12 & 69.67 & 51.63 \\
        MInference & 28.62 & 41.08 & 53.06 & 64.28 & 61.21 & 51.01 & 31.71 & 23.52 & 30.53 & 21.07 & 22.79 & 10.72 & 75.00 & 90.16 & 44.96 & 46.00 & 5.00 & 95.00 & 97.67 & 69.83 & 67.52 & 49.08 \\
        FlexPrefill & 28.07 & 40.96 & 52.41 & 64.07 & 60.72 & 50.91 & 31.65 & 23.97 & 30.02 & 21.62 & 23.86 & 10.88 & 75.50 & 89.37 & 46.38 & 48.50 & 4.50 & 97.00 & 96.00 & 70.74 & 68.93 & 49.34 \\
        XAttention & 27.98 & 42.86 & 52.83 & 63.82 & 60.93 & 53.16 & 33.01 & 23.81 & 31.31 & 21.28 & 22.93 & 10.97 & 73.00 & 91.14 & 45.92 & 47.50 & 3.00 & 97.00 & 99.00 & 71.42 & 69.01 & 49.61 \\
        FlashPrefill & 28.35 & 41.11 & 53.34 & 64.81 & 61.27 & 51.09 & 30.08 & 24.11 & 30.55 & 21.74 & 23.21 & 11.02 & 77.00 & 90.68 & 46.12 & 46.50 & 5.00 & 96.50 & 99.00 & 71.86 & 68.38 & 49.61 \\
        FlashPrefill V2 & 29.74 & 41.58 & 53.82 & 66.80 & 63.66 & 52.62 & 34.11 & 24.94 & 31.00 & 22.04 & 23.45 & 11.13 & 77.00 & 90.99 & 46.59 & 49.00 & 6.50 & 99.50 & 99.50 & 72.51 & 68.91 & \textbf{50.73} \\
        FlashPrefill V2-FP8 & 30.03 & 41.37 & 55.28 & 66.31 & 61.40 & 51.83 & 30.63 & 24.58 & 31.19 & 21.62 & 23.56 & 10.96 & 77.50 & 89.61 & 46.08 & 47.50 & 8.50 & 98.00 & 99.00 & 72.16 & 69.05 & \textbf{50.29} \\
    \bottomrule[1pt]
    \end{tabular}}
    \caption{Performance comparison on all 21 tasks of LongBench~\cite{longbench}, grouped by task category. Results marked with $^*$ (in gray) are measured with direct online FP8 quantization without corrected weights.}
    \label{tab:longbench}
\end{table*}

\section{Experiments}
\label{sec:experiments}
We evaluate FlashPrefill V2 on two widely-recognized long-context benchmarks, RULER~\cite{hsieh2024ruler} and LongBench~\cite{longbench}, across representative LLMs. We first describe the experimental setup, then report accuracy and efficiency results, and finally present ablation studies on the individual components.

\subsection{Setup}
\paragraph{Benchmarks and Models.}RULER~\cite{hsieh2024ruler} stresses long-context retrieval and reasoning at controlled sequence lengths, and LongBench~\cite{longbench} covers a diverse suite of realistic long-context tasks. We evaluate on Llama-3.1-8B-Instruct~\cite{llama3}, Qwen3-4B-Instruct-2507~\cite{qwen3technicalreport}, and Qwen3-30B-A3B-Instruct-2507~\cite{qwen3technicalreport}.

\paragraph{Baselines.}We compare against Full Attention~\cite{attention}, MInference~\cite{minference}, FlexPrefill~\cite{flexprefill}, and XAttention~\cite{xattention}. All methods are evaluated under identical hyperparameter settings on NVIDIA H20 GPUs, the most widely deployed inference GPU in our target deployment environments; all measurements in this section are collected on H20 unless otherwise noted. As dense references, we use FlashAttention-2~\cite{flashattention2} and an FA3/4-aligned dense kernel~\cite{flashattention3,flashattention4} that shares the same warp-specialized pipeline as our sparse operator, so that the measured speedups isolate the benefit of sparsity rather than kernel engineering.

\paragraph{Implementation Details.}The block-sparse operator is compiled with fixed tile sizes: a query tile of $128$ rows (\texttt{kBlockM}) and a key tile of $64$ columns (\texttt{kBlockN}). For all accuracy evaluations, every model shares a single configuration without per-model tuning: selection block size $B=128$, $256$ sink tokens, a local window of $512$ tokens, and $\alpha=0.1$ in the max-based thresholding. All operator-level results are measured on a single H20 GPU. End-to-end results (Tabs.~\ref{tab:ttft},~\ref{tab:serving}, and~\ref{tab:serving_chunk}) are collected from a single SGLang server instance per configuration with tensor parallelism $TP{=}4$ on four H20 GPUs and no data parallelism; the decode attention backend is fixed to FA3/4 across all configurations, and only the prefill attention backend varies. The software stack comprises CUDA 12.9 (driver 535.161.08), PyTorch 2.9.1, CUTLASS 4.3 for the attention kernels, and SGLang 0.5.10~\cite{sglang} for serving.

\begin{table}[!t]
    \centering
    \begin{tabular}{l|cccccc}
    \toprule[1pt]
        Model & 4K & 8K & 16K & 32K & 64K & 128K \\
        \midrule[0.5pt]
        Llama-3.1-8B & 76.0\% & 54.2\% & 33.6\% & 18.7\% & 9.0\% & 4.6\% \\
        Qwen3-4B & 72.0\% & 48.5\% & 29.8\% & 17.6\% & 9.2\% & 4.9\% \\
        Qwen3-30B-A3B & 70.4\% & 46.0\% & 29.6\% & 16.2\% & 9.4\% & 4.9\% \\
    \bottomrule[1pt]
    \end{tabular}
    \caption{Attention density of FlashPrefill V2 measured on needle-in-a-haystack inputs at each sequence length.}
    \label{tab:ruler_density}
\end{table}

\subsection{Accuracy Results}

\begin{table*}[!t]
    \centering
    \resizebox{\textwidth}{!}{
    \begin{tabular}{c|cccccc|cccccc|cccccc}
    \toprule[1pt]
        \multirow{2}{*}{BSZ} & \multicolumn{6}{c|}{FA3/4} & \multicolumn{6}{c|}{FlashPrefill V2} & \multicolumn{6}{c}{FlashPrefill V2-FP8} \\
        \cmidrule(lr){2-7} \cmidrule(lr){8-13} \cmidrule(lr){14-19}
        & 4K & 8K & 16K & 32K & 64K & 128K & 4K & 8K & 16K & 32K & 64K & 128K & 4K & 8K & 16K & 32K & 64K & 128K \\
        \midrule[0.5pt]
        \multicolumn{19}{c}{\textit{Llama-3.1-8B-Instruct}} \\
        \midrule[0.5pt]
        1 & 0.15 & 0.29 & 0.63 & 1.49 & 3.95 & 11.81 & \makecell[c]{0.15\\{\scriptsize(1.02$\times$)}} & \makecell[c]{0.29\\{\scriptsize(1.02$\times$)}} & \makecell[c]{0.58\\{\scriptsize(1.09$\times$)}} & \makecell[c]{1.17\\{\scriptsize(1.28$\times$)}} & \makecell[c]{2.46\\{\scriptsize(1.60$\times$)}} & \makecell[c]{5.49\\{\scriptsize(2.15$\times$)}} & \makecell[c]{0.10\\{\scriptsize(1.53$\times$)}} & \makecell[c]{0.18\\{\scriptsize(1.63$\times$)}} & \makecell[c]{0.35\\{\scriptsize(1.81$\times$)}} & \makecell[c]{0.71\\{\scriptsize(2.09$\times$)}} & \makecell[c]{1.48\\{\scriptsize(2.66$\times$)}} & \makecell[c]{3.23\\{\scriptsize(3.66$\times$)}} \\
        4 & 0.58 & 1.17 & 2.54 & 5.94 & 12.30 & 32.94 & \makecell[c]{0.57\\{\scriptsize(1.01$\times$)}} & \makecell[c]{1.14\\{\scriptsize(1.03$\times$)}} & \makecell[c]{2.32\\{\scriptsize(1.09$\times$)}} & \makecell[c]{4.79\\{\scriptsize(1.24$\times$)}} & \makecell[c]{7.69\\{\scriptsize(1.60$\times$)}} & \makecell[c]{15.39\\{\scriptsize(2.14$\times$)}} & \makecell[c]{0.35\\{\scriptsize(1.66$\times$)}} & \makecell[c]{0.68\\{\scriptsize(1.73$\times$)}} & \makecell[c]{1.36\\{\scriptsize(1.87$\times$)}} & \makecell[c]{2.80\\{\scriptsize(2.12$\times$)}} & \makecell[c]{4.49\\{\scriptsize(2.74$\times$)}} & \makecell[c]{8.84\\{\scriptsize(3.73$\times$)}} \\
        8 & 1.07 & 2.21 & 4.87 & 9.83 & 21.23 & 56.67 & \makecell[c]{1.06\\{\scriptsize(1.01$\times$)}} & \makecell[c]{2.14\\{\scriptsize(1.03$\times$)}} & \makecell[c]{4.39\\{\scriptsize(1.11$\times$)}} & \makecell[c]{7.77\\{\scriptsize(1.27$\times$)}} & \makecell[c]{13.18\\{\scriptsize(1.61$\times$)}} & \makecell[c]{26.13\\{\scriptsize(2.17$\times$)}} & \makecell[c]{0.67\\{\scriptsize(1.58$\times$)}} & \makecell[c]{1.39\\{\scriptsize(1.59$\times$)}} & \makecell[c]{2.81\\{\scriptsize(1.73$\times$)}} & \makecell[c]{4.51\\{\scriptsize(2.18$\times$)}} & \makecell[c]{7.66\\{\scriptsize(2.77$\times$)}} & \makecell[c]{14.93\\{\scriptsize(3.80$\times$)}} \\
        16 & 2.60 & 3.85 & 8.14 & 16.10 & 37.12 & 102.84 & \makecell[c]{2.58\\{\scriptsize(1.01$\times$)}} & \makecell[c]{3.59\\{\scriptsize(1.07$\times$)}} & \makecell[c]{7.31\\{\scriptsize(1.11$\times$)}} & \makecell[c]{12.56\\{\scriptsize(1.28$\times$)}} & \makecell[c]{22.93\\{\scriptsize(1.62$\times$)}} & \makecell[c]{46.96\\{\scriptsize(2.19$\times$)}} & \makecell[c]{1.44\\{\scriptsize(1.80$\times$)}} & \makecell[c]{2.97\\{\scriptsize(1.30$\times$)}} & \makecell[c]{4.62\\{\scriptsize(1.76$\times$)}} & \makecell[c]{7.28\\{\scriptsize(2.21$\times$)}} & \makecell[c]{13.25\\{\scriptsize(2.80$\times$)}} & \makecell[c]{26.82\\{\scriptsize(3.83$\times$)}} \\
        \midrule[0.5pt]
        \multicolumn{19}{c}{\textit{Qwen3-4B-Instruct-2507}} \\
        \midrule[0.5pt]
        1 & 0.09 & 0.19 & 0.43 & 1.13 & 3.35 & 11.10 & \makecell[c]{0.09\\{\scriptsize(1.02$\times$)}} & \makecell[c]{0.18\\{\scriptsize(1.07$\times$)}} & \makecell[c]{0.36\\{\scriptsize(1.19$\times$)}} & \makecell[c]{0.77\\{\scriptsize(1.47$\times$)}} & \makecell[c]{1.71\\{\scriptsize(1.96$\times$)}} & \makecell[c]{4.12\\{\scriptsize(2.70$\times$)}} & \makecell[c]{0.07\\{\scriptsize(1.42$\times$)}} & \makecell[c]{0.12\\{\scriptsize(1.61$\times$)}} & \makecell[c]{0.23\\{\scriptsize(1.87$\times$)}} & \makecell[c]{0.49\\{\scriptsize(2.33$\times$)}} & \makecell[c]{1.08\\{\scriptsize(3.09$\times$)}} & \makecell[c]{2.44\\{\scriptsize(4.55$\times$)}} \\
        4 & 0.35 & 0.74 & 1.69 & 4.35 & 10.61 & 31.47 & \makecell[c]{0.35\\{\scriptsize(1.01$\times$)}} & \makecell[c]{0.69\\{\scriptsize(1.06$\times$)}} & \makecell[c]{1.43\\{\scriptsize(1.18$\times$)}} & \makecell[c]{3.06\\{\scriptsize(1.42$\times$)}} & \makecell[c]{5.38\\{\scriptsize(1.97$\times$)}} & \makecell[c]{11.55\\{\scriptsize(2.72$\times$)}} & \makecell[c]{0.22\\{\scriptsize(1.59$\times$)}} & \makecell[c]{0.43\\{\scriptsize(1.72$\times$)}} & \makecell[c]{0.88\\{\scriptsize(1.93$\times$)}} & \makecell[c]{1.86\\{\scriptsize(2.34$\times$)}} & \makecell[c]{3.26\\{\scriptsize(3.25$\times$)}} & \makecell[c]{6.73\\{\scriptsize(4.67$\times$)}} \\
        8 & 0.64 & 1.38 & 3.25 & 7.40 & 18.25 & 53.84 & \makecell[c]{0.63\\{\scriptsize(1.02$\times$)}} & \makecell[c]{1.29\\{\scriptsize(1.07$\times$)}} & \makecell[c]{2.70\\{\scriptsize(1.20$\times$)}} & \makecell[c]{5.05\\{\scriptsize(1.47$\times$)}} & \makecell[c]{9.19\\{\scriptsize(1.99$\times$)}} & \makecell[c]{19.35\\{\scriptsize(2.78$\times$)}} & \makecell[c]{0.43\\{\scriptsize(1.49$\times$)}} & \makecell[c]{0.87\\{\scriptsize(1.59$\times$)}} & \makecell[c]{1.78\\{\scriptsize(1.82$\times$)}} & \makecell[c]{3.06\\{\scriptsize(2.42$\times$)}} & \makecell[c]{5.51\\{\scriptsize(3.31$\times$)}} & \makecell[c]{11.28\\{\scriptsize(4.77$\times$)}} \\
        16 & 1.07 & 2.31 & 5.53 & 12.20 & 31.73 & 97.26 & \makecell[c]{1.05\\{\scriptsize(1.02$\times$)}} & \makecell[c]{2.15\\{\scriptsize(1.08$\times$)}} & \makecell[c]{4.56\\{\scriptsize(1.21$\times$)}} & \makecell[c]{8.18\\{\scriptsize(1.49$\times$)}} & \makecell[c]{15.84\\{\scriptsize(2.00$\times$)}} & \makecell[c]{34.48\\{\scriptsize(2.82$\times$)}} & \makecell[c]{0.90\\{\scriptsize(1.18$\times$)}} & \makecell[c]{1.83\\{\scriptsize(1.26$\times$)}} & \makecell[c]{2.95\\{\scriptsize(1.87$\times$)}} & \makecell[c]{4.92\\{\scriptsize(2.48$\times$)}} & \makecell[c]{9.43\\{\scriptsize(3.37$\times$)}} & \makecell[c]{20.13\\{\scriptsize(4.83$\times$)}} \\
        \midrule[0.5pt]
        \multicolumn{19}{c}{\textit{Qwen3-30B-A3B-Instruct-2507}} \\
        \midrule[0.5pt]
        1 & 0.10 & 0.19 & 0.45 & 1.26 & 4.00 & 13.78 & \makecell[c]{0.09\\{\scriptsize(1.02$\times$)}} & \makecell[c]{0.18\\{\scriptsize(1.09$\times$)}} & \makecell[c]{0.36\\{\scriptsize(1.26$\times$)}} & \makecell[c]{0.78\\{\scriptsize(1.61$\times$)}} & \makecell[c]{1.75\\{\scriptsize(2.29$\times$)}} & \makecell[c]{4.17\\{\scriptsize(3.30$\times$)}} & \makecell[c]{0.09\\{\scriptsize(1.06$\times$)}} & \makecell[c]{0.15\\{\scriptsize(1.24$\times$)}} & \makecell[c]{0.30\\{\scriptsize(1.50$\times$)}} & \makecell[c]{0.61\\{\scriptsize(2.07$\times$)}} & \makecell[c]{1.32\\{\scriptsize(3.03$\times$)}} & \makecell[c]{3.00\\{\scriptsize(4.59$\times$)}} \\
        4 & 0.26 & 0.73 & 1.76 & 4.86 & 13.05 & 40.44 & \makecell[c]{0.24\\{\scriptsize(1.09$\times$)}} & \makecell[c]{0.67\\{\scriptsize(1.10$\times$)}} & \makecell[c]{1.42\\{\scriptsize(1.24$\times$)}} & \makecell[c]{3.11\\{\scriptsize(1.56$\times$)}} & \makecell[c]{5.70\\{\scriptsize(2.29$\times$)}} & \makecell[c]{12.16\\{\scriptsize(3.33$\times$)}} & \makecell[c]{0.22\\{\scriptsize(1.18$\times$)}} & \makecell[c]{0.57\\{\scriptsize(1.29$\times$)}} & \makecell[c]{1.15\\{\scriptsize(1.53$\times$)}} & \makecell[c]{2.41\\{\scriptsize(2.01$\times$)}} & \makecell[c]{4.21\\{\scriptsize(3.10$\times$)}} & \makecell[c]{8.59\\{\scriptsize(4.71$\times$)}} \\
        8 & 0.64 & 1.35 & 3.36 & 8.47 & 22.37 & 68.56 & \makecell[c]{0.62\\{\scriptsize(1.03$\times$)}} & \makecell[c]{1.22\\{\scriptsize(1.11$\times$)}} & \makecell[c]{2.65\\{\scriptsize(1.27$\times$)}} & \makecell[c]{5.25\\{\scriptsize(1.61$\times$)}} & \makecell[c]{9.69\\{\scriptsize(2.31$\times$)}} & \makecell[c]{20.35\\{\scriptsize(3.37$\times$)}} & \makecell[c]{0.55\\{\scriptsize(1.15$\times$)}} & \makecell[c]{1.02\\{\scriptsize(1.32$\times$)}} & \makecell[c]{2.11\\{\scriptsize(1.59$\times$)}} & \makecell[c]{4.03\\{\scriptsize(2.10$\times$)}} & \makecell[c]{7.13\\{\scriptsize(3.14$\times$)}} & \makecell[c]{14.35\\{\scriptsize(4.78$\times$)}} \\
        16 & 1.16 & 3.14 & 5.90 & 13.98 & 38.50 & 123.23 & \makecell[c]{1.15\\{\scriptsize(1.01$\times$)}} & \makecell[c]{2.92\\{\scriptsize(1.08$\times$)}} & \makecell[c]{4.60\\{\scriptsize(1.28$\times$)}} & \makecell[c]{8.49\\{\scriptsize(1.65$\times$)}} & \makecell[c]{16.55\\{\scriptsize(2.33$\times$)}} & \makecell[c]{36.21\\{\scriptsize(3.40$\times$)}} & \makecell[c]{1.04\\{\scriptsize(1.12$\times$)}} & \makecell[c]{1.73\\{\scriptsize(1.81$\times$)}} & \makecell[c]{3.66\\{\scriptsize(1.61$\times$)}} & \makecell[c]{6.47\\{\scriptsize(2.16$\times$)}} & \makecell[c]{12.16\\{\scriptsize(3.17$\times$)}} & \makecell[c]{25.51\\{\scriptsize(4.83$\times$)}} \\
    \bottomrule[1pt]
    \end{tabular}}
    \caption{End-to-end time-to-first-token (TTFT, seconds) of Llama-3.1-8B-Instruct, Qwen3-4B-Instruct-2507, and Qwen3-30B-A3B-Instruct-2507 served by SGLang on the needle-in-a-haystack workload. Each entry is averaged over $30$ repeated measurements. Parenthesized numbers are speedups over the FA3/4 backend.}
    \label{tab:ttft}
\end{table*}

\paragraph{RULER.}Tab.~\ref{tab:ruler} reports RULER scores at controlled sequence lengths (left) together with the end-to-end attention operator speedups (right). Across all three models, FlashPrefill V2 stays within $1.1$ points of Full attention on average ($87.79$ vs $88.82$ on Llama-3.1-8B, $86.23$ vs $87.06$ on Qwen3-4B, and $91.76$ vs $92.05$ on Qwen3-30B-A3B), and the FP8 variant costs only an additional $0.4$ to $1.2$ points. The gap is largest at 128K where the density drops below $5\%$ (Tab.~\ref{tab:ruler_density} lists the measured densities at each sequence length), yet V2 remains within $1.8$ points of Full attention. Meanwhile, the sparse operator including index selection already outperforms FA2 at 4K and reaches $27.2\times$ to $27.5\times$ at 128K in BF16 and $47.3\times$ to $47.6\times$ in FP8.

\begin{table*}[!t]
    \centering
    \resizebox{\textwidth}{!}{%
    \begin{tabular}{l|l|cc|cc|c|c}
    \toprule[1pt]
        \multirow{2}{*}{\makecell{Rate\\(req/s)}} & \multirow{2}{*}{Backend} & \multicolumn{2}{c|}{TTFT (s)} & \multicolumn{2}{c|}{TPOT (ms)} & \multicolumn{2}{c}{Throughput} \\
         & & P50 & P99 & P50 & P99 & (tok/s) & (req/s) \\
        \midrule[0.5pt]
        \multicolumn{8}{c}{\textit{Qwen3-4B-Instruct-2507}} \\
        \midrule[0.5pt]
        \multirow{3}{*}{1} & FA3/4 & 76.73 & 154.64 & 419 & 1170 & 24.0 & 0.37 \\
         & FlashPrefill V2 & \textbf{17.23 (4.45x)} & \textbf{48.73 (3.17x)} & \textbf{258 (1.62x)} & \textbf{601 (1.94x)} & \textbf{45.5 (1.90x)} & 0.71 \\
         & FlashPrefill V2-FP8 & \textbf{2.79 (27.47x)} & \textbf{17.27 (8.95x)} & \textbf{47 (8.84x)} & \textbf{487 (2.40x)} & \textbf{58.4 (2.44x)} & 0.91 \\
        \midrule[0.5pt]
        \multirow{3}{*}{4} & FA3/4 & 79.05 & 161.54 & 513 & 1102 & 23.8 & 0.37 \\
         & FlashPrefill V2 & \textbf{42.28 (1.87x)} & \textbf{80.34 (2.01x)} & \textbf{258 (1.99x)} & \textbf{509 (2.16x)} & \textbf{47.3 (1.99x)} & 0.74 \\
         & FlashPrefill V2-FP8 & \textbf{24.47 (3.23x)} & \textbf{46.05 (3.51x)} & \textbf{165 (3.10x)} & \textbf{331 (3.33x)} & \textbf{82.3 (3.46x)} & 1.29 \\
        \midrule[0.5pt]
        \multirow{3}{*}{8} & FA3/4 & 79.55 & 161.01 & 463 & 1110 & 23.9 & 0.37 \\
         & FlashPrefill V2 & \textbf{43.83 (1.81x)} & \textbf{80.12 (2.01x)} & \textbf{236 (1.96x)} & \textbf{523 (2.12x)} & \textbf{48.3 (2.03x)} & 0.76 \\
         & FlashPrefill V2-FP8 & \textbf{25.83 (3.08x)} & \textbf{46.24 (3.48x)} & \textbf{152 (3.05x)} & \textbf{302 (3.68x)} & \textbf{82.2 (3.45x)} & 1.28 \\
        \midrule[0.5pt]
        \multirow{3}{*}{16} & FA3/4 & 81.91 & 160.99 & 481 & 1110 & 23.8 & 0.37 \\
         & FlashPrefill V2 & \textbf{43.22 (1.90x)} & \textbf{78.68 (2.05x)} & \textbf{263 (1.83x)} & \textbf{501 (2.21x)} & \textbf{48.9 (2.05x)} & 0.76 \\
         & FlashPrefill V2-FP8 & \textbf{25.14 (3.26x)} & \textbf{45.16 (3.56x)} & \textbf{152 (3.17x)} & \textbf{285 (3.89x)} & \textbf{85.5 (3.59x)} & 1.34 \\
        \midrule[0.5pt]
        \multicolumn{8}{c}{\textit{Qwen3-30B-A3B-Instruct-2507}} \\
        \midrule[0.5pt]
        \multirow{3}{*}{1} & FA3/4 & 90.69 & 195.43 & 582 & 1335 & 19.9 & 0.31 \\
         & FlashPrefill V2 & \textbf{17.63 (5.14x)} & \textbf{52.15 (3.75x)} & \textbf{241 (2.42x)} & \textbf{604 (2.21x)} & \textbf{44.5 (2.24x)} & 0.70 \\
         & FlashPrefill V2-FP8 & \textbf{7.93 (11.44x)} & \textbf{21.92 (8.91x)} & \textbf{102 (5.69x)} & \textbf{637 (2.10x)} & \textbf{56.1 (2.82x)} & 0.88 \\
        \midrule[0.5pt]
        \multirow{3}{*}{4} & FA3/4 & 101.44 & 195.33 & 462 & 1312 & 19.9 & 0.31 \\
         & FlashPrefill V2 & \textbf{46.26 (2.19x)} & \textbf{85.57 (2.28x)} & \textbf{255 (1.81x)} & \textbf{520 (2.52x)} & \textbf{45.5 (2.29x)} & 0.71 \\
         & FlashPrefill V2-FP8 & \textbf{30.62 (3.31x)} & \textbf{57.43 (3.40x)} & \textbf{198 (2.33x)} & \textbf{407 (3.22x)} & \textbf{64.6 (3.25x)} & 1.01 \\
        \midrule[0.5pt]
        \multirow{3}{*}{8} & FA3/4 & 105.52 & 195.34 & 481 & 1333 & 19.9 & 0.31 \\
         & FlashPrefill V2 & \textbf{45.73 (2.31x)} & \textbf{83.04 (2.35x)} & \textbf{234 (2.05x)} & \textbf{529 (2.52x)} & \textbf{47.0 (2.36x)} & 0.73 \\
         & FlashPrefill V2-FP8 & \textbf{33.17 (3.18x)} & \textbf{58.46 (3.34x)} & \textbf{165 (2.92x)} & \textbf{369 (3.62x)} & \textbf{66.9 (3.36x)} & 1.05 \\
        \midrule[0.5pt]
        \multirow{3}{*}{16} & FA3/4 & 99.68 & 195.20 & 547 & 1347 & 19.9 & 0.31 \\
         & FlashPrefill V2 & \textbf{45.20 (2.21x)} & \textbf{81.95 (2.38x)} & \textbf{266 (2.06x)} & \textbf{514 (2.62x)} & \textbf{47.4 (2.38x)} & 0.74 \\
         & FlashPrefill V2-FP8 & \textbf{32.47 (3.07x)} & \textbf{57.33 (3.40x)} & \textbf{185 (2.95x)} & \textbf{359 (3.75x)} & \textbf{67.7 (3.41x)} & 1.06 \\
    \bottomrule[1pt]
    \end{tabular}}
    \caption{Open-loop serving results on Qwen3-4B-Instruct-2507 and Qwen3-30B-A3B-Instruct-2507 under Poisson arrivals with mixed prompt lengths (4K--128K RULER needle-in-a-haystack, 100 requests per cell, 64 output tokens each). Parenthesized numbers are speedups over the FA3/4 backend.}
    \label{tab:serving}
\end{table*}

\paragraph{LongBench.}Tab.~\ref{tab:longbench} reports the per-task results on LongBench. FlashPrefill V2 achieves the highest average among all sparse operators on every model: $49.31$ on Llama-3.1-8B (vs.\ $48.25$ for the strongest baseline, XAttention), $46.96$ on Qwen3-4B (vs.\ $45.90$), and $50.73$ on Qwen3-30B-A3B (vs.\ $49.61$), staying within $0.9$ points of full attention. The advantage is most pronounced on retrieval-sensitive synthetic tasks: on English passage retrieval, V2 retains $96.0$ to $99.5$ points while the baselines drop to $91.0$ to $97.0$, and on the counting task V2 remains the closest to full attention on all three models. Summarization tasks are nearly lossless, with V2 occasionally exceeding full attention (e.g., $44.76$ vs.\ $43.59$ on SAMSum for Llama-3.1-8B). The FP8 variant, even with direct online quantization, still outperforms every baseline on average ($48.76$/$46.07$/$50.29$). Tab.~\ref{tab:ruler_density} lists the attention densities measured on needle-in-a-haystack inputs.

\subsection{Efficiency Results}

The operator-level speedups over FA2 are reported on the right side of Tab.~\ref{tab:ruler}. Fig.~\ref{fig:attention_speedup} further compares against the FA3/4-aligned dense kernel and prior sparse operators at batch size $4$: at 128K context, our operator attains $27.2\times$ (BF16) and $47.3\times$ (FP8) over FA2, $17.5\times$ and $30.5\times$ over the FA3/4-aligned dense kernel, and clearly surpasses prior sparse operators ($18.7\times$ for FlashPrefill, $5.4\times$ for FlexPrefill, $3.4\times$ for XAttention, and $2.8\times$ for MInference). The density stays around $70\%$ at 4K and drops to about $5\%$ at 128K (Tab.~\ref{tab:ruler_density}), which is the source of the growing speedup.

\paragraph{End-to-End Serving Latency.}Tab.~\ref{tab:ttft} reports the end-to-end time-to-first-token of FlashPrefill V2 deployed in SGLang, with batch sizes from $1$ to $16$ under continuous batching. The operator-level gains translate directly into serving latency: at 128K context, FlashPrefill V2 reduces TTFT by $2.1\times$ to $3.4\times$ in BF16 and $3.7\times$ to $4.8\times$ in FP8 across all three models and batch sizes (e.g., from $123.2$\,s to $36.2$\,s in BF16 and $25.5$\,s in FP8 at batch size $16$ on Qwen3-30B-A3B). The speedup is nearly batch-invariant, indicating that the sparse operator scales with continuous batching in the same way as the dense backend. As expected, the end-to-end speedup is bounded by the attention share of total prefill compute and is therefore smaller than the operator-level one; it grows with sequence length as attention becomes dominant, and is largest on Qwen3-30B-A3B, where attention accounts for a larger fraction of prefill compute. At 4K, where the density remains around $70\%$, BF16 stays at parity with FA3/4 ($1.01\times$ to $1.09\times$), where the modest attention savings are largely offset by the index stage overhead, while FP8 already delivers $1.1\times$ to $1.8\times$.

\begin{table*}[!t]
    \centering
    \resizebox{\textwidth}{!}{%
    \begin{tabular}{l|l|cc|cc|c|c}
    \toprule[1pt]
        \multirow{2}{*}{\makecell{Chunk\\size}} & \multirow{2}{*}{Backend} & \multicolumn{2}{c|}{TTFT (s)} & \multicolumn{2}{c|}{TPOT (ms)} & \multicolumn{2}{c}{Throughput} \\
         & & P50 & P99 & P50 & P99 & (tok/s) & (req/s) \\
        \midrule[0.5pt]
        \multicolumn{8}{c}{\textit{Qwen3-4B-Instruct-2507}} \\
        \midrule[0.5pt]
        \multirow{3}{*}{8K} & FA3/4 & 80.98 & 165.14 & 597 & 1228 & 22.5 & 0.35 \\
         & FlashPrefill V2 & \textbf{50.15 (1.61x)} & \textbf{95.23 (1.73x)} & \textbf{360 (1.66x)} & \textbf{684 (1.80x)} & \textbf{38.7 (1.72x)} & 0.60 \\
         & FlashPrefill V2-FP8 & \textbf{33.47 (2.42x)} & \textbf{61.20 (2.70x)} & \textbf{259 (2.30x)} & \textbf{529 (2.32x)} & \textbf{60.7 (2.70x)} & 0.95 \\
        \midrule[0.5pt]
        \multirow{3}{*}{16K} & FA3/4 & 79.42 & 161.96 & 580 & 1203 & 22.9 & 0.36 \\
         & FlashPrefill V2 & \textbf{45.09 (1.76x)} & \textbf{85.47 (1.89x)} & \textbf{315 (1.84x)} & \textbf{615 (1.96x)} & \textbf{43.9 (1.91x)} & 0.69 \\
         & FlashPrefill V2-FP8 & \textbf{26.80 (2.96x)} & \textbf{50.34 (3.22x)} & \textbf{193 (3.00x)} & \textbf{382 (3.15x)} & \textbf{73.3 (3.20x)} & 1.15 \\
        \midrule[0.5pt]
        \multicolumn{8}{c}{\textit{Qwen3-30B-A3B-Instruct-2507}} \\
        \midrule[0.5pt]
        \multirow{3}{*}{8K} & FA3/4 & 96.61 & 199.39 & 693 & 1494 & 18.7 & 0.29 \\
         & FlashPrefill V2 & \textbf{79.81 (1.21x)} & \textbf{149.52 (1.33x)} & \textbf{574 (1.21x)} & \textbf{1200 (1.24x)} & \textbf{24.6 (1.32x)} & 0.38 \\
         & FlashPrefill V2-FP8 & \textbf{63.23 (1.53x)} & \textbf{121.53 (1.64x)} & \textbf{505 (1.37x)} & \textbf{1009 (1.48x)} & \textbf{29.2 (1.56x)} & 0.46 \\
        \midrule[0.5pt]
        \multirow{3}{*}{16K} & FA3/4 & 94.78 & 195.84 & 672 & 1466 & 19.0 & 0.30 \\
         & FlashPrefill V2 & \textbf{46.47 (2.04x)} & \textbf{88.15 (2.22x)} & \textbf{323 (2.08x)} & \textbf{652 (2.25x)} & \textbf{42.1 (2.21x)} & 0.66 \\
         & FlashPrefill V2-FP8 & \textbf{33.72 (2.81x)} & \textbf{62.19 (3.15x)} & \textbf{250 (2.69x)} & \textbf{464 (3.16x)} & \textbf{57.1 (3.00x)} & 0.89 \\
    \bottomrule[1pt]
    \end{tabular}}
    \caption{Open-loop serving at $16$ req/s with chunked prefill at chunk sizes of $8$K and $16$K (same workload as Tab.~\ref{tab:serving}, where chunking is disabled). Parenthesized numbers are speedups over the FA3/4 backend at the same chunk size.}
    \label{tab:serving_chunk}
\end{table*}

\paragraph{Open-Loop Serving.}Tab.~\ref{tab:serving} moves from fixed batches to an open-loop setting: requests arrive as a Poisson process at $1$ to $16$ req/s with prompt lengths mixed over 4K--128K, so prefills and decodes of different requests continuously share the engine. The FA3/4 backend is saturated at every rate (P50 TTFT of $77$--$106$\,s, request throughput capped at $0.31$--$0.37$ req/s), whereas FlashPrefill V2 sustains about twice the request throughput ($0.70$--$0.76$ req/s) and cuts the P50 TTFT to $17$--$46$\,s; with FP8 the request throughput rises to $0.88$--$1.34$ req/s and the P50 TTFT drops further to $2.8$--$33$\,s. The speedup is largest at the lowest rate ($4.5\times$--$5.1\times$ P50 at $1$ req/s in BF16 and $11.4\times$--$27.5\times$ in FP8), where faster prefills shorten queueing much more than at higher rates, and remains $1.8\times$--$2.3\times$ in BF16 and $3.1\times$--$3.3\times$ in FP8 once the system is loaded. Shorter prefills also benefit decoding: because long dense prefills block the decode steps of concurrently running requests, P50 TPOT improves by $1.6\times$--$2.4\times$ in BF16 and $2.3\times$--$8.8\times$ in FP8, and output throughput rises by $1.9\times$--$2.4\times$ and $2.4\times$--$3.6\times$ respectively.

\paragraph{Compatibility with Chunked Prefill.}Chunked prefill~\cite{sarathiserve}, which bounds the number of tokens processed per scheduling step, is the standard mechanism in production engines for protecting inter-token latency from long prefills. Tab.~\ref{tab:serving_chunk} repeats the open-loop experiment at $16$ req/s with chunk sizes of $8$K and $16$K. Chunking barely affects the dense backend's TTFT, but it erodes our speedup because the index selection is re-run on every chunk and the mandatory tail blocks raise the effective density of short chunks: the P50 TTFT speedup in BF16 drops from $1.9\times$ without chunking to $1.6\times$ at $8$K on Qwen3-4B, and from $2.2\times$ to $1.2\times$ on Qwen3-30B-A3B. Even so, FlashPrefill V2 remains faster than the dense backend at $8$K ($1.2\times$--$1.6\times$ in BF16 and $1.5\times$--$2.4\times$ in FP8), and a $16$K chunk recovers most of the margin ($1.8\times$--$2.0\times$ and $2.8\times$--$3.0\times$ respectively), so we recommend chunk sizes of at least $8$K when deploying FlashPrefill V2 together with chunked prefill.

\begin{figure*}[!t]
    \centering
    \begin{minipage}[t]{0.49\textwidth}
        \centering
        \includegraphics[width=\linewidth]{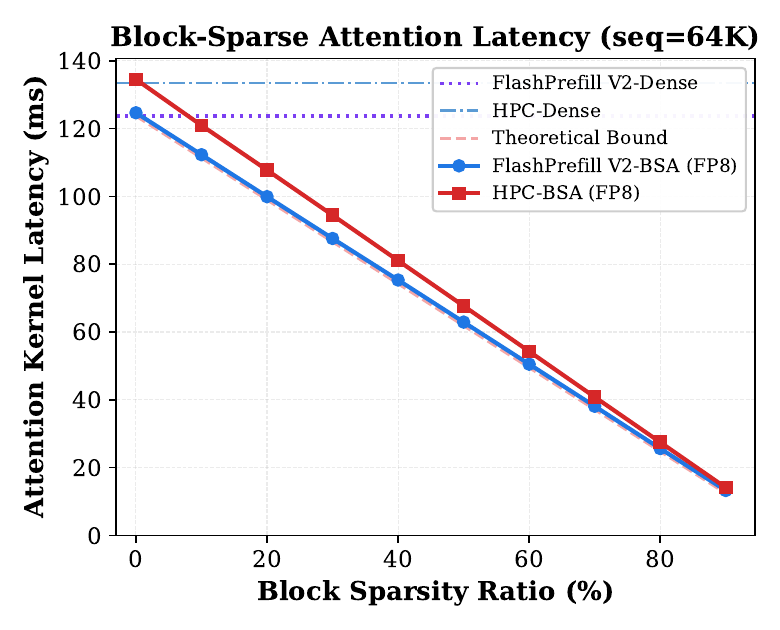}
        \captionof{figure}{Block-sparse attention latency at 64K sequence length in FP8. HPC-BSA does not support BF16, so only FP8 is compared. Both sparse operators track the theoretical bound; FlashPrefill V2 is $6$ to $7\%$ faster than HPC-BSA at every sparsity level.}
        \label{fig:bsa_compare}
    \end{minipage}
    \hfill
    \begin{minipage}[t]{0.49\textwidth}
        \centering
        \includegraphics[width=\linewidth]{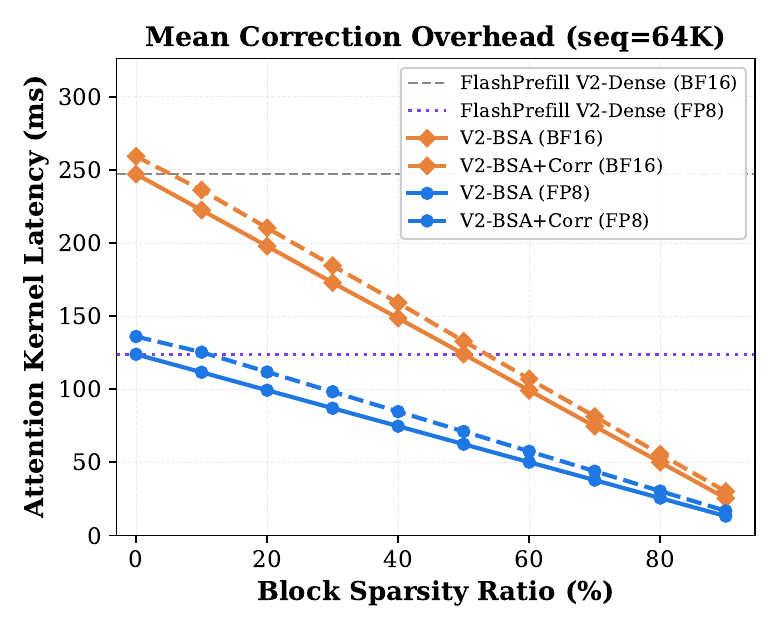}
        \captionof{figure}{Overhead of mean correction at 64K sequence length. The corrected variants (dashed) stay close to the uncorrected ones (solid) in both BF16 and FP8, and the absolute overhead shrinks as sparsity increases.}
        \label{fig:bsa_corr}
    \end{minipage}
\end{figure*}

\paragraph{Comparison with the Production Kernel.}
Fig.~\ref{fig:bsa_compare} compares our block-sparse operator against the block-sparse attention (BSA) kernel in HPC-Ops~\cite{hpcops}, a production inference operator library. Since the HPC-Ops BSA kernel only supports FP8, the comparison is conducted in FP8 at 64K sequence length. Our operator is consistently faster than HPC-BSA across all sparsity levels (by $6$ to $7\%$), and both dense references are comparable (123.7 ms vs 133.4 ms), confirming that the gain comes from the sparse execution path rather than the dense backbone. Tab.~\ref{tab:bsa_design} summarizes the design differences behind this gap.

\paragraph{Overhead of Mean Correction.}
Fig.~\ref{fig:bsa_corr} quantifies the runtime cost of the correction term at 64K. Because all pruned blocks within a query tile are replaced by a single pass over their pooled K/V means, the extra latency is small and nearly constant in absolute terms: at most $12.2$ ms in the dense limit and only $3.6$ to $4.5$ ms at $90\%$ sparsity. The relative overhead therefore stays within $5\%$ to $14\%$ up to $50\%$ sparsity, and reaches $18\%$ (BF16) to $27\%$ (FP8) only at $90\%$ sparsity, the regime where the correction is most needed for accuracy.

\begin{table}[!t]
    \centering
    \resizebox{\columnwidth}{!}{
    \begin{tabular}{l|c|c}
    \toprule[1pt]
        & \textbf{HPC-Ops BSA} & \textbf{FlashPrefill V2} \\
        \midrule[0.5pt]
        Sparse index format & dense mask, padded, per Q head & CSR of selected blocks, per KV head \\
        Index memory @ 64K / 128K & 8\,MB / 32\,MB (any density) & 3\,MB / 6\,MB (9\% / 5\% density) \\
        GEMM--softmax overlap & serial & intra-warpgroup pingpong \\
        Paged KV addressing & per-page TMA (page size $\mid$ 128) & cp.async, arbitrary page size \\
        KV splitting & none & count-balanced CSR splits \\
        Mean correction & none & fused into the mainloop \\
        Precision & FP8 only & BF16 + FP8 \\
    \bottomrule[1pt]
    \end{tabular}}
    \caption{Design comparison with the HPC-Ops block-sparse attention kernel. Index memory is evaluated at batch size $1$ with $32$ query heads and $4$ KV heads: the dense mask stores one byte per (query head, $128$-token Q tile, $128$-token KV tile) regardless of density, whereas the CSR index stores only the selected $64$-token tiles as int32 at the measured needle-in-a-haystack densities (Tab.~\ref{tab:ruler_density}), is organized per KV head under PackGQA, and matches the SGLang index format. The fused correction path has no counterpart in HPC-Ops BSA.}
    \label{tab:bsa_design}
\end{table}

\subsection{Ablation Study}
\paragraph{Effect of Mean Correction.}
Tab.~\ref{tab:ablation_correction} ablates the mean correction term (Sec.~\ref{sec:correction}) on RULER with Qwen3-4B-Instruct-2507, comparing the full FlashPrefill V2 pipeline against a variant in which all unselected blocks are simply discarded. Without correction, the dropped probability mass grows as the density falls, so the accuracy gap widens with sequence length: in BF16 the variant trails the corrected pipeline by only $0.3$ points on average up to 32K, but by $0.8$ points at 64K and $0.9$ points at 128K, where fewer than $5\%$ of the blocks survive selection (Tab.~\ref{tab:ruler_density}). Restoring the correction term recovers nearly all of this degradation at a negligible runtime cost (Fig.~\ref{fig:bsa_corr}), keeping FlashPrefill V2 within $0.9$ points of full attention on average and within $1.5$ points even at 128K.
The correction becomes substantially more important in FP8: without it, the FP8 variant loses $2.3$ points on average and $6.2$ points at 128K relative to its corrected counterpart, compared with only $0.5$ and $0.9$ points in BF16. We attribute this to the compressed score margins under quantization, which let the blocks discarded by the max-based threshold carry a larger share of the softmax mass, so recovering them matters more. With correction enabled, the FP8 pipeline tracks BF16 closely ($85.82$ vs.\ $86.23$ on average), confirming that the correction remains effective even when the pooled statistics are quantized.

\begin{table}[!t]
    \centering
    \begin{tabular}{l|cccccc|c}
    \toprule[1pt]
        Method & 4K & 8K & 16K & 32K & 64K & 128K & Avg. \\
        \midrule[0.5pt]
        Full Attention & 94.82 & 92.98 & 91.68 & 88.86 & 82.81 & 71.22 & 87.06 \\
        V2 w/o correction & 94.38 & 92.06 & 91.18 & 87.21 & 80.92 & 68.86 & 85.77 \\
        FlashPrefill V2 & 94.49 & 92.42 & 91.56 & 87.48 & 81.67 & 69.76 & 86.23 \\
        \midrule[0.5pt]
        V2-FP8 w/o correction & 93.78 & 91.82 & 90.68 & 85.04 & 76.86 & 62.78 & 83.49 \\
        FlashPrefill V2-FP8 & 94.13 & 92.37 & 91.36 & 87.47 & 80.59 & 68.97 & 85.82 \\
    \bottomrule[1pt]
    \end{tabular}
    \caption{Ablation of the mean correction term on RULER with Qwen3-4B-Instruct-2507, in both BF16 and FP8. Discarding unselected blocks without compensation degrades accuracy increasingly toward longer contexts, while the zero-order correction stays close to full attention at all lengths in both precisions.}
    \label{tab:ablation_correction}
\end{table}

\paragraph{Effect of the Selection Threshold.}
Tab.~\ref{tab:ablation_alpha} sweeps the threshold $\alpha$ of the max-based selection on RULER at 64K with Qwen3-4B-Instruct-2507 in FP8, and reports the measured attention density together with the scores with and without mean correction. As $\alpha$ decreases from $0.2$ to $0.0125$, the density rises from $5.2\%$ to $23.6\%$. The corrected FP8 pipeline is remarkably flat across the entire sweep, staying within $2.7$ points of full attention ($82.81$) even at $5.2\%$ density and within $2.3$ points at the default $\alpha=0.1$, whereas removing the correction consistently costs $2.7$ to $5.4$ further points, with the gap widening as the density falls. Mean correction therefore enlarges the usable sparsity range: with compensation, $\alpha$ can be pushed to $0.2$ at 64K---halving the density relative to the default---while retaining $80.12$ points.

\begin{table}[!t]
    \centering
    \resizebox{\linewidth}{!}{
    \begin{tabular}{l|ccccc}
    \toprule[1pt]
        $\alpha$ & 0.2 & 0.1 & 0.05 & 0.025 & 0.0125 \\
        \midrule[0.5pt]
        Density (\%) & 5.2 & 9.2 & 14.1 & 19.8 & 23.6 \\
        V2-FP8 w/o correction & 74.68\textcolor{red}{($-$5.44)} & 76.86\textcolor{red}{($-$3.73)} & 77.16\textcolor{red}{($-$3.51)} & 77.41\textcolor{red}{($-$3.31)} & 78.02\textcolor{red}{($-$2.73)} \\
        FlashPrefill V2-FP8 & 80.12 & 80.59 & 80.67 & 80.72 & 80.75 \\
    \bottomrule[1pt]
    \end{tabular}}
    \caption{Ablation of the selection threshold $\alpha$ on RULER at 64K with Qwen3-4B-Instruct-2507 in FP8. Density is the fraction of selected attention blocks. \textcolor{red}{Red numbers} are the score drops of the uncorrected variant relative to the corrected pipeline at the same $\alpha$. Full attention scores $82.81$.}
    \label{tab:ablation_alpha}
\end{table}

\section{Conclusion}
In this paper, we present FlashPrefill V2, which evolves FlashPrefill from an algorithmic prototype toward practical long-context prefilling. A zero-order mean correction term compensates pruned blocks with their pooled K/V statistics inside the softmax computation, confining the accuracy loss to within about one point on RULER and LongBench averages, and to within $1.8$ points at 128K on RULER where fewer than $5\%$ of the blocks are computed, while proving particularly effective under FP8 quantization. The redesigned sparse operator aligns with the FlashAttention-3/4 execution model---PackGQA memory access, warp-specialized producer-consumer pipelines, intra-warpgroup pingpong overlap, and FP8 execution---attaining up to $\mathbf{27.19\times}$ (BF16) and $\mathbf{47.26\times}$ (FP8) speedups over FlashAttention-2 at 128K, and a $\mathbf{30.49\times}$ speedup over an FA3/4-aligned dense baseline. With native support for paged KV cache and continuous batching, FlashPrefill V2 integrates into SGLang as a standard attention backend without any model-side changes, reducing end-to-end time-to-first-token by up to $3.4\times$ (BF16) and $4.8\times$ (FP8) at 128K under continuous batching. We hope this work helps bring sparse attention from algorithmic exploration toward practical serving.

\bibliography{main}

@article{llama3,
  title={The Llama 3 Herd of Models},
  author={Grattafiori, Aaron and Dubey, Abhimanyu and Jauhri, Abhinav and Pandey, Abhinav and others},
  journal={arXiv preprint arXiv:2407.21783},
  year={2024}
}

@article{llama,
  title   = {Llama: Open and efficient foundation language models},
  author  = {Touvron, Hugo and Lavril, Thibaut and Izacard, Gautier and Martinet, Xavier and Lachaux, Marie-Anne and Lacroix, Timoth{\'e}e and Rozi{\`e}re, Baptiste and Goyal, Naman and Hambro, Eric and Azhar, Faisal and others},
  journal = {arXiv preprint arXiv:2302.13971},
  year    = {2023}
}

@article{llama2,
  title   = {Llama 2: Open foundation and fine-tuned chat models},
  author  = {Touvron, Hugo and Martin, Louis and Stone, Kevin and Albert, Peter and Almahairi, Amjad and Babaei, Yasmine and Bashlykov, Nikolay and Batra, Soumya and Bhargava, Prajjwal and Bhosale, Shruti and others},
  journal = {arXiv preprint arXiv:2307.09288},
  year    = {2023}
}

@article{attention,
  title   = {Attention is all you need},
  author  = {Vaswani, Ashish and Shazeer, Noam and Parmar, Niki and Uszkoreit, Jakob and Jones, Llion and Gomez, Aidan N and Kaiser, {\L}ukasz and Polosukhin, Illia},
  journal = {Advances in Neural Information Processing Systems},
  volume  = {30},
  year    = {2017}
}

@inproceedings{minference,
  author = {Huiqiang Jiang and Yucheng Li and Chengruidong Zhang and Qianhui Wu and Xufang Luo and Surin Ahn and Zhenhua Han and Amir H. Abdi and Dongsheng Li and Chin-Yew Lin and Yuqing Yang and Lili Qiu},
  booktitle = {Advances in Neural Information Processing Systems},
  title = {{MI}nference 1.0: Accelerating Pre-filling for Long-Context {LLM}s via Dynamic Sparse Attention},
  year = {2024}
}

@inproceedings{flexprefill,
    title={FlexPrefill: A Context-Aware Sparse Attention Mechanism for Efficient Long-Sequence Inference},
    author={Xunhao Lai and Jianqiao Lu and Yao Luo and Yiyuan Ma and Xun Zhou},
    booktitle={International Conference on Learning Representations},
    year={2025}
}

@inproceedings{xattention,
  title     = {XAttention: Block Sparse Attention with Antidiagonal Scoring},
  author    = {Xu, Ruyi and Xiao, Guangxuan and Huang, Haofeng and Guo, Junxian and Han, Song},
  booktitle = {International Conference on Machine Learning},
  year      = {2025}
}

@article{qwen3technicalreport,
      title={Qwen3 Technical Report}, 
      author={An Yang and Anfeng Li and Baosong Yang and Beichen Zhang and Binyuan Hui and Bo Zheng and Bowen Yu and Chang Gao and Chengen Huang and Chenxu Lv and Chujie Zheng and Dayiheng Liu and Fan Zhou and Fei Huang and Feng Hu and Hao Ge and Haoran Wei and Huan Lin and Jialong Tang and Jian Yang and Jianhong Tu and Jianwei Zhang and Jianxin Yang and Jiaxi Yang and Jing Zhou and Jingren Zhou and Junyang Lin and Kai Dang and Keqin Bao and Kexin Yang and Le Yu and Lianghao Deng and Mei Li and Mingfeng Xue and Mingze Li and Pei Zhang and Peng Wang and Qin Zhu and Rui Men and Ruize Gao and Shixuan Liu and Shuang Luo and Tianhao Li and Tianyi Tang and Wenbiao Yin and Xingzhang Ren and Xinyu Wang and Xinyu Zhang and Xuancheng Ren and Yang Fan and Yang Su and Yichang Zhang and Yinger Zhang and Yu Wan and Yuqiong Liu and Zekun Wang and Zeyu Cui and Zhenru Zhang and Zhipeng Zhou and Zihan Qiu},
      journal={arXiv preprint arXiv:2505.09388},
      year={2025},
      eprint={2505.09388},
      archivePrefix={arXiv},
      primaryClass={cs.CL},
      url={https://arxiv.org/abs/2505.09388}
}

@article{mobamixtureblockattention,
      title={MoBA: Mixture of Block Attention for Long-Context LLMs}, 
      author={Enzhe Lu and Zhejun Jiang and Jingyuan Liu and Yulun Du and Tao Jiang and Chao Hong and Shaowei Liu and Weiran He and Enming Yuan and Yuzhi Wang and Zhiqi Huang and Huan Yuan and Suting Xu and Xinran Xu and Guokun Lai and Yanru Chen and Huabin Zheng and Junjie Yan and Jianlin Su and Yuxin Wu and Neo Y. Zhang and Zhilin Yang and Xinyu Zhou and Mingxing Zhang and Jiezhong Qiu},
      journal={arXiv preprint arXiv:2502.13189},
      year={2025},
      eprint={2502.13189},
      archivePrefix={arXiv},
      primaryClass={cs.LG},
      url={https://arxiv.org/abs/2502.13189}
}

@article{optimizingmixtureblockattention,
      title={Optimizing Mixture of Block Attention}, 
      author={Guangxuan Xiao and Junxian Guo and Kasra Mazaheri and Song Han},
      journal={arXiv preprint arXiv:2511.11571},
      year={2025},
      eprint={2511.11571},
      archivePrefix={arXiv},
      primaryClass={cs.CL},
      url={https://arxiv.org/abs/2511.11571}
}

@inproceedings{native-sparse-attention,
    title = "Native Sparse Attention: Hardware-Aligned and Natively Trainable Sparse Attention",
    author = "Yuan, Jingyang  and
      Gao, Huazuo  and
      Dai, Damai  and
      Luo, Junyu  and
      Zhao, Liang  and
      Zhang, Zhengyan  and
      Xie, Zhenda  and
      Wei, Yuxing  and
      Wang, Lean  and
      Xiao, Zhiping  and
      Wang, Yuqing  and
      Ruan, Chong  and
      Zhang, Ming  and
      Liang, Wenfeng  and
      Zeng, Wangding",
    booktitle = "Annual Meeting of the Association for Computational Linguistics",
    year = "2025"
}

@article{zhao2025infllmv2,
  title={InfLLM-V2: Dense-Sparse Switchable Attention for Seamless Short-to-Long Adaptation},
  author={Weilin Zhao and Zihan Zhou and Zhou Su and Chaojun Xiao and Yuxuan Li and Yanghao Li and Yudi Zhang and Weilun Zhao and Zhen Li and Yuxiang Huang and Ao Sun and Xu Han and Zhiyuan Liu},
  journal={arXiv preprint arXiv:2509.24663},
  year={2025},
  eprint={2509.24663},
  archivePrefix={arXiv},
  primaryClass={cs.CL},
  url={https://arxiv.org/abs/2509.24663}
}

@inproceedings{zhang2023h2o,
  title={H2O: Heavy-Hitter Oracle for Efficient Generative Inference of Large Language Models},
  author={Zhang, Zhenyu and Sheng, Ying and others},
  booktitle={Advances in Neural Information Processing Systems},
  year={2023}
}

@article{beltagy2020longformer,
  title={Longformer: The Long-Document Transformer},
  author={Beltagy, Iz and Peters, Matthew E and Cohan, Arman},
  journal={arXiv preprint arXiv:2004.05150},
  year={2020},
  eprint={2004.05150},
  archivePrefix={arXiv},
  primaryClass={cs.CL},
  url={https://arxiv.org/abs/2004.05150}
}

@article{hsieh2024ruler,
  title={RULER: What's the Real Context Size of Your Long-Context Language Models?},
  author={Cheng-Ping Hsieh and Simeng Sun and Samuel Kriman and Shantanu Acharya and Dima Rekesh and Fei Jia and Yang Zhang and Boris Ginsburg},
  year={2024},
  journal={arXiv preprint arXiv:2404.06654},
  eprint={2404.06654},
  archivePrefix={arXiv},
  primaryClass={cs.CL},
  url={https://arxiv.org/abs/2404.06654}
}

@article{flashattention2,
      title={FlashAttention-2: Faster Attention with Better Parallelism and Work Partitioning}, 
      author={Tri Dao},
      year={2023},
        journal={arXiv preprint arXiv:2307.08691},
      eprint={2307.08691},
      archivePrefix={arXiv},
      primaryClass={cs.CL},
      url={https://arxiv.org/abs/2307.08691}
}

@article{flashprefillv1,
  title   = {FlashPrefill: Instantaneous Pattern Discovery and Thresholding for Ultra-Fast Long-Context Prefilling},
  author  = {Fan, Qihang and Huang, Huaibo and Wu, Zhiying and Wang, Juqiu and Wang, Bingning and He, Ran},
  journal = {arXiv preprint arXiv:2603.06199},
  year    = {2026}
}

@inproceedings{vllm,
  title     = {Efficient Memory Management for Large Language Model Serving with PagedAttention},
  author    = {Kwon, Woosuk and Li, Zhuohan and Zhuang, Siyuan and Sheng, Ying and Zheng, Lianmin and Yu, Cody Hao and Gonzalez, Joseph and Zhang, Hao and Stoica, Ion},
  booktitle = {Proceedings of the 29th Symposium on Operating Systems Principles (SOSP)},
  year      = {2023}
}

@inproceedings{sglang,
  title     = {SGLang: Efficient Execution of Structured Language Model Programs},
  author    = {Zheng, Lianmin and Yin, Liangsheng and Xie, Zhiqiang and Sun, Chuyue and Huang, Jeff and Yu, Cody Hao and Cao, Shiyi and Kozyrakis, Christos and Stoica, Ion and Gonzalez, Joseph E. and Barrett, Clark and Sheng, Ying},
  booktitle = {Advances in Neural Information Processing Systems},
  year      = {2024}
}

@article{flashattention3,
  title   = {FlashAttention-3: Fast and Accurate Attention with Asynchrony and Low-precision},
  author  = {Shah, Jay and Bikshandi, Ganesh and Zhang, Ying and Thakkar, Vijay and Ramade, Pradeep and Dao, Tri},
  journal = {arXiv preprint arXiv:2407.08608},
  year    = {2024}
}

@article{flashattention4,
  title   = {FlashAttention-4: Algorithm and Kernel Pipelining Co-Design for Asymmetric Hardware Scaling},
  author  = {Zadouri, Ted and Hoehnerbach, Markus and Shah, Jay and Liu, Timmy and Thakkar, Vijay and Dao, Tri},
  journal = {arXiv preprint arXiv:2603.05451},
  year    = {2026}
}

@article{solattn,
  title   = {Sol-Attn: Accelerating Video Generation Inference via On-the-Fly Attention Sparsification},
  author  = {Li, Haopeng and Li, Yitong and Chen, Junsong and Ye, Tian and Liu, Haozhe and Yu, Jincheng and Wang, Duomin and Zhang, Ruihua and Xie, Zeke and Xie, Enze and Han, Song},
  journal = {arXiv preprint arXiv:2607.24027},
  year    = {2026}
}

@inproceedings{longbench,
  title     = {LongBench: A Bilingual, Multitask Benchmark for Long Context Understanding},
  author    = {Bai, Yushi and Lv, Xin and Zhang, Jiajie and Lyu, Hongchang and Tang, Jiankai and Huang, Zhidian and Du, Zhengxiao and Liu, Xiao and Zeng, Aohan and Hou, Lei and Dong, Yuxiao and Tang, Jie and Li, Juanzi},
  booktitle = {Proceedings of the 62nd Annual Meeting of the Association for Computational Linguistics (ACL)},
  year      = {2024}
}

@misc{hpcops,
  title         = {{HPC-Ops}: High Performance Computing Operators for Large Language Model Inference},
  author        = {{Tencent Hunyuan AI Infra Team}},
  year          = {2026},
  howpublished  = {\url{https://github.com/Tencent/hpc-ops}},
  note          = {Accessed: 2026-08-17}
}

@article{uniprefill,
  title  = {UniPrefill: Universal Long-Context Prefill Acceleration via Block-wise Dynamic Sparsification},
  author = {Fan, Qihang and Huang, Huaibo and Wu, Zhiying and Wang, Bingning and He, Ran},
  journal= {arXiv preprint arXiv:2605.06221},
  year   = {2026}
}

@inproceedings{mala,
  title     = {Rectifying Magnitude Neglect in Linear Attention},
  author    = {Fan, Qihang and Huang, Huaibo and Ai, Yuang and He, Ran},
  booktitle = {Proceedings of the IEEE/CVF International Conference on Computer Vision (ICCV)},
  year      = {2025}
}

@article{streamingllm,
  title  = {Efficient Streaming Language Models with Attention Sinks},
  author = {Xiao, Guangxuan and Tian, Yuandong and Chen, Beidi and Han, Song and Lewis, Mike},
  journal= {arXiv preprint arXiv:2309.17453},
  year   = {2023}
}

@article{snapkv,
  title  = {SnapKV: LLM Knows What You are Looking for Before Generation},
  author = {Li, Yuhong and Huang, Yingbing and Yang, Bowen and Venkitesh, Bharat and Locatelli, Acyr and Ye, Hanchen and Cai, Tianle and Lewis, Patrick and Chen, Deming},
  journal= {arXiv preprint arXiv:2404.14469},
  year   = {2024}
}

@article{pyramidkv,
  title  = {PyramidKV: Dynamic KV Cache Compression based on Pyramidal Information Funneling},
  author = {Cai, Zefan and Zhang, Yichi and Gao, Bofei and Liu, Yuliang and Li, Yucheng and Liu, Tianyu and Lu, Keming and Xiong, Wayne and Dong, Yue and Hu, Junjie and Xiao, Wen},
  journal= {arXiv preprint arXiv:2406.02069},
  year   = {2024}
}

@article{bigbird,
  title  = {Big Bird: Transformers for Longer Sequences},
  author = {Zaheer, Manzil and Guruganesh, Guru and Dubey, Avinava and Ainslie, Joshua and Alberti, Chris and Ontanon, Santiago and Pham, Philip and Ravula, Anirudh and Wang, Qifan and Yang, Li and Ahmed, Amr},
  journal= {arXiv preprint arXiv:2007.14062},
  year   = {2020}
}

@article{reformer,
  title  = {Reformer: The Efficient Transformer},
  author = {Kitaev, Nikita and Kaiser, {\L}ukasz and Levskaya, Anselm},
  journal= {arXiv preprint arXiv:2001.04451},
  year   = {2020}
}

@article{vyas2020clustered,
  title  = {Fast Transformers with Clustered Attention},
  author = {Vyas, Apoorv and Katharopoulos, Angelos and Fleuret, Fran{\c{c}}ois},
  journal= {arXiv preprint arXiv:2007.04825},
  year   = {2020}
}

@article{infllm,
  title  = {InfLLM: Training-Free Long-Context Extrapolation for LLMs with an Efficient Context Memory},
  author = {Xiao, Chaojun and Zhang, Pengle and Han, Xu and Xiao, Guangxuan and Lin, Yankai and Zhang, Zhengyan and Liu, Zhiyuan and Sun, Maosong},
  journal= {arXiv preprint arXiv:2402.04617},
  year   = {2024}
}

@article{quest,
  title  = {Quest: Query-Aware Sparsity for Efficient Long-Context LLM Inference},
  author = {Tang, Jiaming and Zhao, Yilong and Zhu, Kan and Xiao, Guangxuan and Kasikci, Baris and Han, Song},
  journal= {arXiv preprint arXiv:2406.10774},
  year   = {2024}
}

@article{sparq,
  title  = {SparQ Attention: Bandwidth-Efficient LLM Inference},
  author = {Ribar, Luka and Chelombiev, Ivan and Hudlass-Galley, Luke and Blake, Charlie and Luschi, Carlo and Orr, Douglas},
  journal= {arXiv preprint arXiv:2312.04985},
  year   = {2023}
}

@article{tactic,
  title  = {Tactic: Adaptive Sparse Attention with Clustering and Distribution Fitting for Long-Context LLMs},
  author = {Zhu, Kan and Tang, Tian and Xu, Qinyu and Gu, Yile and Zeng, Zhichen and Kadekodi, Rohan and Zhao, Liangyu and Li, Ang and Krishnamurthy, Arvind and Kasikci, Baris},
  journal= {arXiv preprint arXiv:2502.12216},
  year   = {2025}
}

@article{duoattention,
  title  = {DuoAttention: Efficient Long-Context LLM Inference with Retrieval and Streaming Heads},
  author = {Xiao, Guangxuan and Tang, Jiaming and Zuo, Jingwei and Guo, Junxian and Yang, Shang and Tang, Haotian and Fu, Yao and Han, Song},
  journal= {arXiv preprint arXiv:2410.10819},
  year   = {2024}
}

@article{triforce,
  title  = {TriForce: Lossless Acceleration of Long Sequence Generation with Hierarchical Speculative Decoding},
  author = {Sun, Hanshi and Chen, Zhuoming and Yang, Xinyu and Tian, Yuandong and Chen, Beidi},
  journal= {arXiv preprint arXiv:2404.11912},
  year   = {2024}
}

@article{mminference,
  title  = {MMInference: Accelerating Pre-filling for Long-Context VLMs via Modality-Aware Permutation Sparse Attention},
  author = {Li, Yucheng and Jiang, Huiqiang and Zhang, Chengruidong and Wu, Qianhui and Luo, Xufang and Ahn, Surin and Abdi, Amir H. and Li, Dongsheng and Gao, Jianfeng and Yang, Yuqing and Qiu, Lili},
  journal= {arXiv preprint arXiv:2504.16083},
  year   = {2025}
}

@article{retnet,
  title  = {Retentive Network: A Successor to Transformer for Large Language Models},
  author = {Sun, Yutao and Dong, Li and Huang, Shaohan and Ma, Shuming and Xia, Yuqing and Xue, Jilong and Wang, Jianyong and Wei, Furu},
  journal= {arXiv preprint arXiv:2307.08621},
  year   = {2023}
}

@article{rwkv7,
  title  = {RWKV-7 "Goose" with Expressive Dynamic State Evolution},
  author = {Peng, Bo and Zhang, Ruichong and Goldstein, Daniel and Alcaide, Eric and Du, Xingjian and Hou, Haowen and Lin, Jiaju and Liu, Jiaxing and Lu, Janna and Merrill, William and Song, Guangyu and Tan, Kaifeng and Utpala, Saiteja and Wilce, Nathan and Wind, Johan S. and Wu, Tianyi and Wuttke, Daniel and Zhou-Zheng, Christian},
  journal= {arXiv preprint arXiv:2503.14456},
  year   = {2025}
}

@article{distserve,
  title  = {DistServe: Disaggregating Prefill and Decoding for Goodput-optimized Large Language Model Serving},
  author = {Zhong, Yinmin and Liu, Shengyu and Chen, Junda and Hu, Jianbo and Zhu, Yibo and Liu, Xuanzhe and Jin, Xin and Zhang, Hao},
  journal= {arXiv preprint arXiv:2401.09670},
  year   = {2024}
}

@article{mooncake,
  title  = {Mooncake: A KVCache-centric Disaggregated Architecture for LLM Serving},
  author = {Qin, Ruoyu and Li, Zheming and He, Weiran and Zhang, Mingxing and Wu, Yongwei and Zheng, Weimin and Xu, Xinran},
  journal= {arXiv preprint arXiv:2407.00079},
  year   = {2024}
}

@inproceedings{sarathiserve,
  title     = {Taming Throughput-Latency Tradeoff in LLM Inference with Sarathi-Serve},
  author    = {Agrawal, Amey and Kedia, Nitin and Panwar, Ashish and Mohan, Jayashree and Kwatra, Nipun and Gulavani, Bhargav and Tumanov, Alexey and Ramjee, Ramachandran},
  booktitle = {18th USENIX Symposium on Operating Systems Design and Implementation (OSDI)},
  year      = {2024}
}

\end{CJK*}
\end{document}